\pdfoutput=1
\documentclass[11pt]{article}

\newcommand{\rmv}[1]{}

\usepackage{bm}
\usepackage{amsmath} 
\usepackage{amssymb}
\usepackage{amsfonts}
\usepackage{multirow}
\usepackage{tablefootnote}
\usepackage{array}
\usepackage{subfig}
\usepackage{booktabs}
\usepackage{graphicx}
\usepackage{algorithm}
\usepackage{url}
\usepackage{tabularx}
\usepackage{xcolor}

\newcolumntype{L}[1]{>{\raggedright\let\newline\\\arraybackslash\hspace{0pt}}m{#1}}
\newcolumntype{C}[1]{>{\centering\let\newline\\\arraybackslash\hspace{0pt}}m{#1}}
\newcolumntype{R}[1]{>{\raggedleft\let\newline\\\arraybackslash\hspace{0pt}}m{#1}}

\usepackage[final]{acl}

\usepackage{times}
\usepackage{latexsym}

\usepackage[T1]{fontenc}
\usepackage[utf8]{inputenc}

\usepackage{microtype}

\newcommand{\modelname}{SPIN}

\title{SPIN: Sparsifying and Integrating Internal Neurons \\in Large Language Models for Text Classification}

\usepackage{fdsymbol}

\author{
    Difan Jiao$^\clubsuit$\thanks{\quad Equal contribution.}\quad Yilun Liu$^\vardiamondsuit$\footnotemark[1] \\
    \textbf{Zhenwei Tang$^\clubsuit$\quad Daniel Matter$^\vardiamondsuit$\quad Jürgen Pfeffer$^\vardiamondsuit$\quad Ashton Anderson$^\clubsuit$} \\
    $^\clubsuit$University of Toronto, Canada\quad 
    $^\vardiamondsuit$Technical University of Munich, Germany\\
    \small\texttt{difanjiao@cs.toronto.edu\quad yilun.liu@tum.de\quad josephtang@cs.toronto.edu} \\
    \small\texttt{\{daniel.matter, juergen.pfeffer\}@tum.de\quad ashton@cs.toronto.edu}
}

\begin{document}
\maketitle

\begin{abstract}
Among the many tasks that Large Language Models (LLMs) have revolutionized is text classification. 
Current text classification paradigms, however, rely solely on the output of the final layer in the LLM, with the rich information contained in internal neurons largely untapped. 
In this study, we present \modelname{}\footnote{Code repository and interactive web demo are publicly available via \url{https://github.com/difanj0713/SPIN} and \url{https://liuyilun2000.github.io/spin-visualization/}}: a model-agnostic framework that sparsifies and integrates internal neurons of intermediate layers of LLMs for text classification. 
Specifically, \modelname{} sparsifies internal neurons by linear probing-based salient neuron selection layer by layer, avoiding noise from unrelated neurons and ensuring efficiency. 
The cross-layer salient neurons are then integrated to serve as multi-layered features for the classification head.
Extensive experimental results show that our proposed framework can significantly improve text classification accuracy, efficiency, and interpretability. 

\rmv{
Since LLMs progress from lower-level, simple concepts to higher-level, complex concepts across their layered architecture, the internal neurons of LLMs contain rich information.
However, this potential is largely untapped, with the classification process typically relying solely on the output of the final layer.} 

\end{abstract}

\begin{figure*}
    \centering
    % \vspace{-0.1in}
    \includegraphics[width=1\textwidth]{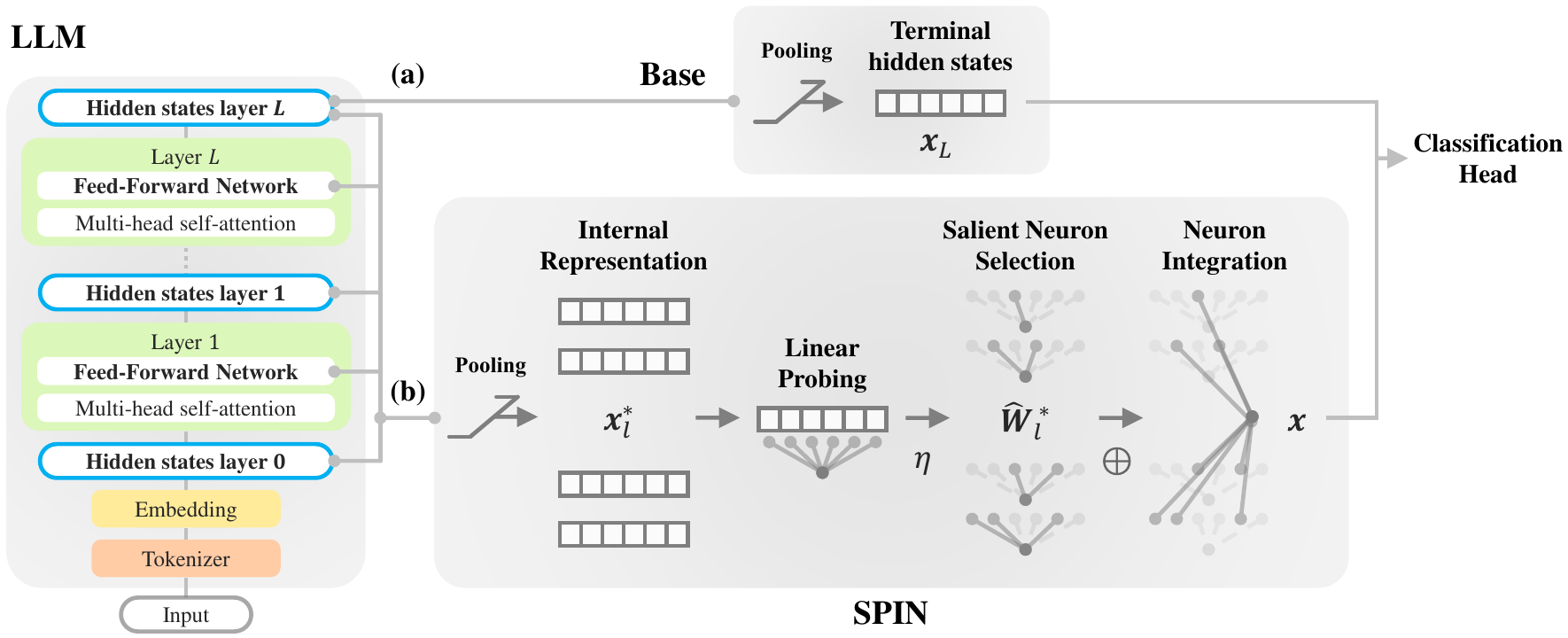}
    \caption{Overview of (a) baseline method that only uses the terminal hidden states; (b) SPIN that uses \textit{sparsified} and \textit{integrated} internal neurons from each intermediate layers to feed the classification head.}
    \label{fig:arch}
    \vspace{-0.1in}
\end{figure*}

\section{Introduction}

%current landscape
Large Language Models (LLMs) have achieved state-of-the-art performance in a wide spectrum of important tasks, including text classification such as sentiment analysis \citep{srivastava2022beyond}. 
Although prompting methods~\cite{wei2022chain,kojima2022large} have gained popularity in deploying LLMs for text classification, employing a classification head with these models remains a dominant paradigm, mainly due to its superior performance in specific tasks~\cite{chang2023survey}.

%problem - "however"
This prevailing paradigm directly uses the terminal hidden states from models that are either pretrained on general tasks or fine-tuned for specific tasks. However, it is fundamentally limited in several ways. First, the implicit internal structures that contribute to LLMs' impressive performance is neglected, forgoing potential performance gains. Second, achieving competitive performance often necessitates fine-tuning LLMs for the task at hand, which in turn can be computationally expensive. Third, this approach inherently lacks interpretability, since it treats models as black boxes. As the demand grows for models that not only perform well but are also interpretable and cost-efficient to train and run, moving beyond the current paradigm is becoming increasingly important.

%insights as motivation
We have reason to believe that delving into the internal of LLMs would bear fruit. As recent studies in AI interpretability~\citep{radford2017learning, bills2023language, gurnee2023language} have revealed, internal representations of artificial neural networks are remarkably adept at capturing essential features, yet the full potential of these insights in the realm of text classification awaits further exploration and demonstration.

%Solution
In this work, we introduce \modelname{}: a model-agnostic framework that \underline{sp}arsifies and \underline{in}tegrates internal neurons of intermediate layers of LLMs for text classification. As shown in Figure \ref{fig:arch}, instead of relying solely on the final layer's hidden states, our method uses internal representations (feed-forward network activations and hidden states) as multi-layered features to enhance the classification head.
These raw internal representations require further processing before being utilized, as internal neurons do not contribute equally to predictions. Irrelevant neurons that introduce noise and extraneous information can be counterproductive, potentially diluting the impact of crucial features. Therefore, \modelname{} employs a linear probing based method to select salient neurons layer by layer, effectively sparsifying the internal representation.
The selected neurons are then integrated across layer to serve as curated multi-layer features for text classification, ensuring that the textual features encompass a full spectrum from lower-level, simpler concepts to higher-level, more complex understandings across the hierarchical learning architecture of LLMs. 

\modelname{} presents multiple advantages that distinguish it in the realm of text classification with LLMs. Primarily, its model-agnostic nature allows it to be employed upon various LLMs as a plug-and-play component. Also, the use of curated internal representations as features enables \modelname{} to outperform conventional methods that rely solely on the terminal outputs.
When applied on pretrained models, \modelname{} can achieve performance comparable to state-of-the-art baseline methods that involve fine-tuning LLMs, accomplished by significantly improved training efficiency. This is achieved by requiring only forward passes with LLM weights untouched, and limiting trainable parameters to probing and the classification head, making \modelname{} a cost-effective alternative to fine-tuning for text classification.
In terms of inference efficiency, \modelname{} enables early exiting, with up to 99\% of the performance preserved from processing only 60\% of LLM layers, significantly speeding up the inference process.
Additionally, with its white-box approach of linear probing on internal neurons, \modelname{} enhances both intrinsic and post-hoc interpretability.

Our main contributions are summarized as:
\begin{itemize}
    \item We propose \modelname{}, a model-agnostic text classification framework  that leverages sparsified and integrated internal neurons from intermediate layers of LLMs, moving beyond conventional reliance on terminal hidden states;
    \item We conduct extensive experiments to demonstrate \modelname{}'s superior performance, improved efficiency in training and inference, and enhanced intrinsic and post-hoc interpretability in text classification.
\end{itemize}

\section{Methodology}
\subsection{Overview}
% Joseph Draft
In contrast to conventional text classification methods that rely exclusively on the output from the final layer of LLMs (as in Figure~\ref{fig:arch} (a)), \modelname{} utilizes internal representations from intermediate layers of LLMs for text classification. 
As shown in Figure~\ref{fig:arch} (b), \modelname{} first sparsifies internal neurons with linear probing-based salient neuron selection to exclude noise from unrelated neurons and enhance efficiency.
The cross-layer salient neurons are then integrated to serve as multi-grained features for the classification head.

\subsection{Neuron Sparsification}
\label{sec:sparsify}
% Joseph Draft
The internal representations from all layers of a LLM can be obtained in a single forward pass. However, these raw representations require further processing before they can be effectively utilized for integrated multi-grained text classification. This necessity arises from the fact that not every internal neuron contributes uniformly to text classification tasks within a particular domain. Additionally, unrelated neurons can be detrimental because they may introduce noise and unnecessary information, which could potentially weaken crucial features. Consequently, we pinpoint and select neurons that exhibit the highest salience and utility for the targeted task.

\paragraph{LLM Internal Representations.}
We first extract layer-wise internal representations from LLMs:
\begin{equation}
    \bm{x}_l = \text{Extract}(\text{LLM} | \bm{s})\in\mathbb{R}^{L\times D}, 
\end{equation}
where the internal representation of the $l^{th}$ layer $\bm{x}_l$ is obtained by extracting the hidden states, i.e., the output of each transformer layer with dimension $D_{\text{hs}}$, or the activations, i.e., intermediate representation within the feed-forward network of each transformer layer with dimension $D_{\text{act}}$, of the LLM given the input sentence $\bm{s}$ with length $L$.
Hidden states and activations are extensively utilized as internal representations in interpretability research, as evidenced by various studies~\citep{durrani2020analyzing, burns2022discovering, gurnee2023language} for hidden states, and~\citep{bills2023language, gurnee2023finding} for activations. We thus treat the selection of the internal representation as a hyperparameter, i.e., $D \in \{D_{\text{hs}}, D_{\text{act}}\}$.
The pooling operation is subsequently applied to ensure fixed-dimension internal representations for sentences of variant lengths.
\begin{equation}
    \bm{x}_l^* = \text{Pooling}(\bm{x}_l) \in\mathbb{R}^{D},
\end{equation}
where $\bm{x}_l^*$ denotes the extracted and pooled internal representation.

\paragraph{Linear Probing.} 
% Joseph Draft
% \textcolor{orange}{TODO: LR+L1 used, why LR --> linear representation hypothesis + feature importance, why L1 --> encourage sparse features}
\label{linear-probing}
We apply linear probing~\citep{alain2016understanding} to identify the salient neurons in each layer of LLMs for the targeted task. This approach is well-established for interpretability studies in LLMs~\citep{dalvi2019one, suau2020finding, wang2022finding, gurnee2023finding}, which employs a simple linear model to interpret the saliency of neurons within neural networks by training on their frozen internal representations for specific tasks.
Linear probing suits our needs for two key reasons.
First, its effectiveness is supported by the \emph{linear representation hypothesis}, that neural network features are linearly represented~\citep{mikolov2013efficient, mikolov2013linguistic, elhage2022softmax}. This suggests that linear models are sufficiently complex to capture the nuanced relationships within internal neurons.
Second, its inherent simplicity ensures that our focus remains on frozen internal representations as task-relevant features, rather than on learning additional task-specific dynamics upon them. This facilitates our subsequent salient neuron selection process. 

Specifically, we use the frozen internal representation of each layer $\bm{x}_{l,i}^*$ as features of input sentence $\bm{s}_i$ to train layer-wise linear models for the targeted task: 
\begin{equation}
\resizebox{\hsize}{!}{$
    \min_{\bm{W}_l, \bm{b}_l} \quad \frac{1}{N} \sum_{i=1}^{N} \mathcal{L}(y_i, \sigma(\bm{W}_l \bm{x}_{l, i}^* + \bm{b}_l)) + \lambda \sum_{j} ||w_{l,j}||_1,
$}
\end{equation}
where $\sigma(\cdot)$, $\lambda$, and $N$ represent the sigmoid function, the regularization coefficient, and the number of training sentences, respectively. The weights and bias of the linear model $\bm{W}_l$ and $\bm{b}_l$ are learned by optimizing the linear model with the binary cross-entropy loss $\mathcal{L}$ with label $y_i$.
By using $\mathcal{L}^1$-regularized (Lasso) logistic regression, i.e., adding $\lambda \sum_{j}||w_{l,j}||_1$, the magnitude of the learned weights $\bm{W}_l$ can be interpreted as indicators of the relative importance or contribution of each neuron to the prediction \citep{guyon2003introduction, ng2004feature}.
In particular, larger weights signify a greater influence on the model's output, thereby marking those neurons as particularly salient for the targeted task. 
Additionally, the use of Lasso logistic regression encourages the sparsity of model weights, thereby enhancing the distinction between salient and non-salient neurons \citep{tibshirani1996regression}.

\paragraph{Salient Neuron Selection.}
Then we gather the identified salient neurons. 
The learned weights are first normalized to enhance fair selection.
\begin{equation}
    \hat{w}_{l,i} = \frac{\|w_{l,i}\|}{\sum_{j=1}^{|\bm{W}_l|} \|w_{l,j}\|}, \quad i = 1, 2, \ldots, |\bm{W}_l|,
\end{equation}
where \(\hat{w}_{l,i}\) denotes the normalized weight of the \(i^{th}\) element in the linear probing weights \(\bm{W}_l\). 
Following normalization, the selection of salient neurons is guided by the sparsification threshold $\eta$, which serves as a metric for determining the cumulative contribution of the most significant neurons.
Specifically, we select the largest \(\hat{w}_{l,i}\) until their cumulative summation reaches the $\eta$. This step involves sorting \(\hat{w}_{l,i}\) in descending order to obtain \(\bm{\hat{W}}_l^\downarrow\) and identifying the smallest subset \(\bm{\hat{W}}_l^* \subseteq \bm{\hat{W}}_l^\downarrow\) whose cumulative sum is at least \(\eta\):
\begin{equation}
    \sum_{\hat{w}_{l,i} \in \bm{\hat{W}}_l^*} \hat{w}_{l,i} \geq \eta.
\end{equation}
The internal neurons are then sparsified based on their saliency, identified as follows:
\begin{equation}
    \bm{N}_l = \{i \mid \hat{w}_{l,i} \in \bm{\hat{W}}_l^*\},
\end{equation}
where $\bm{N}_l$ denotes the positions of the salient neurons in layer $l$, highlighting those most relevant for the text classification task based on their normalized weights.

\subsection{Neuron Integration}
% Joseph Draft
\label{agg}
LLMs exhibit a hierarchical learning structure that they transit from encoding lower-level, simpler concepts to capturing higher-level, more complex understandings across their layered architecture, and the internal neurons inherently encapsulate a wealth of information.
Following the sparsification of internal neurons from each respective layer, we proceed to integrate them as cross-layer multi-grained representations into the classification head:
\begin{equation}
    \min_{\bm{W}, \bm{b}} \quad \frac{1}{N} \sum_{i=1}^{N} \mathcal{L}(y_i, \sigma(\bm{W} \bm{x}_i + \bm{b})),
\label{eq:clf_head}
\end{equation}
Conventional LLM-based text classifiers regard the pooled hidden states of the final layer as frozen features of the $i^{th}$ input sentence $\bm{s}_i$, i.e., the textual features $\bm{x}_i = \bm{x}_{L, i}^*$, where $L$ denotes the total number of stacked layers in the LLM. 
The final layer, while representing the LLM's cumulative understanding into a single output, might overlook or underutilize the nuanced and specialized knowledge encoded in the internal neurons of intermediate layers.
Therefore, \modelname{} employs the cross-layer integrated representation as multi-grained features for classifying the $i^{th}$ input sentence $\bm{s}_i$:
\begin{equation}
    \bm{x}_i = \bigoplus_l^{L}\{\bm{x}_{l,i,j}^* | j \in \bm{N}_l\},
\end{equation}
where $\bigoplus$ denotes the concatenation operation.

\begin{table*}[t]
\small
    \centering
  	\renewcommand\tabcolsep{5pt}
    \begin{tabular}{l|C{3em}C{2.5em}R{3em}|C{3em}C{2.5em}R{3em}|C{3em}C{2.5em}R{3em}}
    \toprule
         & \multicolumn{3}{c|}{\textbf{IMDb (Acc.)}} & \multicolumn{3}{c|}{\textbf{SST-2 (Acc.)}} & \multicolumn{3}{c}{\textbf{EDOS (Macro F1)}}  \\
        \cmidrule(l{3pt}r{3pt}){2-4} \cmidrule(l{3pt}r{3pt}){5-7} \cmidrule(l{3pt}){8-10}
        & Base & \modelname{} & \%impr. & Base & \modelname{} & \%impr. & Base & \modelname{} & \%impr. \\
    \midrule
        DistilBERT  & 86.95 & 89.78 & +3.25 & 81.88 & 83.94 & +2.52 & 65.09 & 75.79 & +16.44 \\
        RoBERTa     & 89.67 & 93.61 & +4.39 & 84.06 & 90.59 & +7.77 & 68.81 & 73.50 & +6.82 \\
        GPT2        & 87.72 & 91.94 & +4.81 & 85.89 & 87.73 & +2.14 & 68.57 & 76.08 & +10.95 \\
        GPT2-M      & 88.59 & 93.92 & +6.02 & 86.12 & 90.25 & +4.80 & 71.17 & 75.74 & +6.42 \\
%        GPT2-L      & 91.75 & 94.28 & +2.76 & 90.14 & 91.97 & +2.03 & 72.05 & 76.33 & +5.94 \\
        GPT2-XL     & 91.86 & 94.92 & +3.33 & 90.02 & 93.23 & +3.57 & 72.56 & 76.79 & +5.83 \\
        Flan-T5-S   & 84.08 \rmv{/77.58} & 91.15 & \textbf{+8.41} & 77.17 & 88.99 & +15.32 & 59.62 & 74.51 & \textbf{+24.97} \\
        Flan-T5     & 90.01 \rmv{79.66/} & 94.14 & +4.59 & 78.26 & 92.32 & \textbf{+17.97} & 66.64 & 78.04 & +17.11 \\
%        Flan-T5-L  & 85.82 & 95.40 & \textbf{+11.16} & 77.41 & 94.84 & \textbf{+22.52} & 65.14 & 79.00 & +21.28 \\
        Flan-T5-XL & 90.50 & \textbf{96.12} & +6.21 & 84.75 & \textbf{95.64} & +12.85 & 70.08 & \textbf{81.48} & +16.27 \\
        \midrule
            SoTA        & \multicolumn{3}{c|}{96.21} & \multicolumn{3}{c|}{97.50} & \multicolumn{3}{c}{82.35} \\
        \bottomrule
    \end{tabular}
    \caption{Performance of \modelname{} and baseline method (Base) over pretrained LLMs, with the state-of-the-art fine-tuned model performance (SoTA) for each dataset. \%impr. denotes percentages of improvement. The best results (except for SoTA) and the largest \%impr. are in boldface.}
    \label{tab:performance-pretrained}
    \vspace{-0.1in}
\end{table*}

\section{Experiments}

We conduct extensive experiments to evaluate our proposed \modelname{} framework for text classification, focusing on three key dimensions: performance, efficiency, and interpretability.

\subsection{Experimental Setup}
% Joseph Draft
\paragraph{Datasets.}
We use three well-established benchmark datasets for text classification.
Namely \href{https://huggingface.co/datasets/imdb}{IMDb}, a widely used movie review dataset for binary sentiment classification,
\href{https://huggingface.co/datasets/sst2}{SST-2}, a fine-grained sentiment analysis dataset with binary labels of the sentiment polarity,
and \href{https://github.com/rewire-online/edos}{EDOS}, a unique dataset with scenarios and outcomes related to ethical dilemmas, labeled by sentiment toward the ethicality of the outcomes. 
For IMDb and EDOS, we use the provided dataset splits for training, validation, and testing.
Whereas SST-2 does not provide ground truth labels of the test set, we thus randomly select 20\% data from the training set for validation and use their provided validation set for testing.

More details and the statistics of datasets can be found in Appendix~\ref{appendix-dataset} and Table~\ref{tab:dataset}. A discussion about implementing \modelname{} for multiclass classification tasks is detailed in Appendix \ref{multiclass}.

\paragraph{Language Models.}
We consider representative pre-trained LLMs of mainstream architectures, including encoder-based models (BERT variants), decoder-based models (GPT-2 variants), and encoder-decoder models (T5 variants). As contemporary LLMs continue to scale in size, we demonstrate \modelname{}'s scalability on LLaMA2-7B and 13B in Appendix \ref{scalability}.

\paragraph{Baselines.}
As shown in Eq.(\ref{eq:clf_head}), a classification head on top of the frozen terminal hidden states is trained for text classification\footnote{For the detailed implementations, we refer to \url{https://github.com/huggingface/transformers/tree/v4.37.2/src/transformers/models}}. In particular, encoder-based and decoder-based methods commonly employ the hidden states associated with the \verb+[CLS]+ token and the final token as the frozen textual features, respectively.
For encoder-decoder models, 
the encoder is optimized for understanding and representing input text, unlike decoders, which are designed to generate text based on the encoded representations. For text classification, where the goal is to understand input text rather than generate new text, the encoder's final layer is naturally the most relevant source of features.
Therefore, the classification head is built on the pooled hidden states of each encoder-decoder model's final encoder layer. 
In the subsequent sections, we refer to these baseline methods as \textit{Base}. 

It is important to note that the baseline classification head undergoes supervised training with the true labels from the datasets, distinguishing it from zero or few-shot prompting methods~\cite{wei2022chain,kojima2022large}.

\paragraph{Implementation Details.}
% hyperparameter tuning
We conduct a grid search to optimize the hyperparameter settings of \modelname{}, including the Lasso regularization coefficient $\lambda$, the sparsification threshold $\eta$, the choice of pooling strategy, and the choice of internal representations (hidden states or FFN activations, as described in Section~\ref{sec:sparsify}). The ranges for the grid searching and the optimal hyperparameter settings can be found in Table~\ref{tab:hyper_space} and Table~\ref{tab:best_hyper} in Appendix \ref{hyperparams}.
% evaluation metric
For performance comparisons, we adhere to the official evaluation metrics specified by each benchmark dataset. In particular, we employ accuracy as the metric for IMDb and SST-2, and Macro-F1 for evaluations on EDOS.

\subsection{Results and Analysis}
\subsubsection{\modelname{} Performance}
\paragraph{Performance Improvement.}
We develop \modelname{} on top of various mainstream model architectures, including encoder-based models (BERT variants), decoder-based models (GPT-2 variants), and encoder-decoder models (T5 variants), demonstrating \modelname{} is compatible with a wide spectrum of pre-trained LLMs. 
As the results shown in Table~\ref{tab:performance-pretrained}, each version of \modelname{} consistently outperforms the corresponding baseline method across all benchmark datasets. Specifically, the performance improvement is as much as 25\% on EDOS compared to the baseline method on Flan-T5-S. Even for larger and more advanced models that are already performing well, \modelname{} can still be a highly effective plug-and-play component to boost the performance, e.g., Flan-T5-XL on IMDb improved by 6\% with \modelname{}.
Therefore, as long as the model weights and architecture are provided, it is promising to use \modelname{} as a model-agnostic component for further performance improvement.

\subsubsection{\modelname{} and Fine-tuning}
In this section, we briefly discuss the comparability and compatibility of \modelname{} with full fine-tuning LLMs for text classification tasks. For the analysis of SPIN in conjunction with Parameter-Efficient Fine-Tuning (PEFT) techniques, a detailed discussion is provided in Appendix \ref{PEFT}. 
\paragraph{Comparability.}

\begin{figure}[t!]
    \centering
    \includegraphics[width=0.45\textwidth]
    {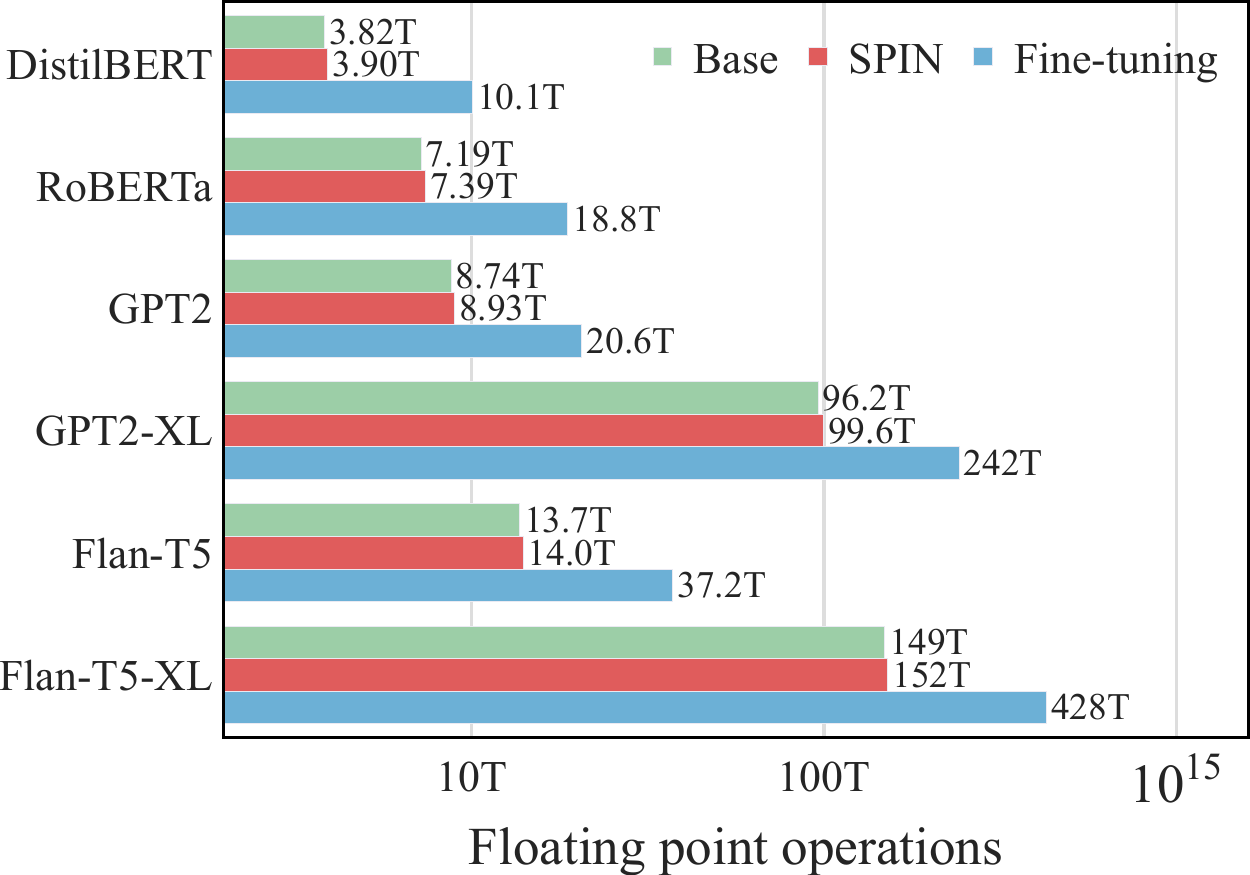}
    \caption{Floating point operations cost for training of baseline, \modelname{}, and full fine-tuning on different models. The cost of \modelname{} is estimated on FFN activations with $\eta=0.5$, and the cost of fine-tuning is estimated based on the lowest demand assumption of 1 epoch.}
    \label{fig:flo}
    \vspace{-0.1in}
\end{figure}

We compare \modelname{} with the state-of-the-art fine-tuned models on XLNet~\cite{yang2019xlnet} for IMDb, T5-11B~\cite{raffel2020exploring} for SST-2, and DeBERTa-v3-base~\cite{he2021debertav3} for EDOS.
As shown in Table~\ref{tab:performance-pretrained}, the best performing \modelname{} upon the considered models can approximate the SoTA performance, e.g., \modelname{} with Flan-T5-XL achieves 99.91\% of the performance of XLNet fine-tuned on IMDb. 

In particular, \modelname{} can adapt to the targeted task without back-propagation and parameter updates on pre-trained LLMs. It utilizes frozen LLM weights and limits the parameter learning process to the classification head, thus having significantly fewer trainable parameters, as discussed in Appendix \ref{param}. This results in reduced storage requirements for hosting the model and potentially less data needed to effectively achieve optimal learning under the scaling law \citep{brown2020language, hoffmann2022training}.

\modelname{}'s better training efficiency is also evidenced by Figure \ref{fig:flo}, where we compare the theoretical computational cost within the training phase of \modelname{} and fine-tuning with floating-point operations \citep{kaplan2020scaling}, with detailed estimation in Appendix \ref{flo}.  
This reduced need for extensive parameter updates, i.e., \modelname{} only requires slightly more computations than forward passes, ensures \modelname{}'s scalability and accessibility, making it a viable option for applications with limited computational resources.
% In Table \ref{tab:performance-pretrained}, we present the test performance of model classification head and \modelname{} for all selected models across all three datasets. 

\begin{table}[t]
    \small
    \centering
    \begin{tabular}{@{\hspace{4pt}}l@{\hspace{4pt}}|@{\hspace{4pt}}c@{\hspace{6pt}}c@{\hspace{4pt}}|@{\hspace{4pt}}c@{\hspace{6pt}}c@{\hspace{4pt}}|@{\hspace{4pt}}c@{\hspace{6pt}}c@{\hspace{4pt}}}
    \toprule
         & \multicolumn{2}{@{\hspace{0pt}}c@{\hspace{4pt}}|@{\hspace{4pt}}}{\textbf{IMDb}} & \multicolumn{2}{@{\hspace{0pt}}c@{\hspace{4pt}}|@{\hspace{4pt}}}{\textbf{SST-2}} & \multicolumn{2}{@{\hspace{0pt}}c@{\hspace{4pt}}}{\textbf{EDOS}}  \\
        \cmidrule(l{-3pt}r){2-3} \cmidrule(l{-3pt}r){4-5} \cmidrule(l{-3pt}){6-7}
         & Base & \modelname{} & Base & \modelname{} & Base & \modelname{} \\
    \midrule
        DistilBERT  & 92.80 & \textbf{92.88} & 91.05 & \textbf{91.19} & 78.74 & \textbf{81.12} \\
        RoBERTa     & 94.67 & \textbf{95.68} & 94.03 & \textbf{94.38} & 80.48 & \textbf{80.88} \\ 
        GPT2        & 94.06 & \textbf{94.50} & 91.51 & \textbf{92.32} & \textemdash & \textemdash \\
%        GPT2-M      & 90.70 & \textbf{94.92} & 91.98 & \textbf{92.32} & \textemdash & \textemdash \\
%        GPT2-L      & 92.74 & \textbf{94.76} & \textemdash & \textemdash & \textemdash & \textemdash \\
%        GPT2-XL     & 93.11 & \textbf{95.12} & \textemdash & \textemdash & \textemdash & \textemdash \\
%        XLNet       & 95.70 & \textbf{95.72} & \textemdash & \textemdash & \textemdash & \textemdash\\
    \bottomrule
    \end{tabular}
    \caption{Performance of \modelname{} and baseline method (Base) over published fine-tuned LLMs. GPT2 fine-tuned on EDOS is not publicly available.}
    \label{tab:performance-finetuned}
\end{table}

\begin{table}[!t]
\centering
\small
\begin{tabular}{lccccc}
\toprule
    & \textbf{20\%} & \textbf{40\%} & \textbf{60\%} & \textbf{80\%} & \textbf{100\%}  \\
\midrule 
    DistilBERT & 85.20 & 84.73 & 87.45 & 88.67 & 89.78 \\
    RoBERTa    & 87.06 & 89.72 & 93.13 & 93.50 & 93.61 \\
    GPT2       & 87.51 & 89.00 & 91.10 & 91.88 & 91.94 \\
    GPT2-M     & 88.52 & 91.36 & 93.36 & 93.92 & 93.92 \\
%    GPT2-L     & 88.72 & 91.47 & 93.99 & 94.05 & 94.28 \\
    GPT2-XL    & 89.66 & 93.15 & 94.73 & 94.92 & 94.92 \\
    Flan-T5-S  & 82.88 & 87.74 & 90.93 & 91.32 & 91.32\\
    Flan-T5    & 84.58 & 92.55 & 94.14 & 94.14 & 94.14\\
%    Flan-T5-L  & 87.77 & 93.88 & 95.40 & 95.40 & 95.40 \\
    Flan-T5-XL & 89.21 & 95.28 & 96.12 & 96.12 & 96.12\\
\bottomrule
\end{tabular}
\caption{Performance of \modelname{} on IMDb with early-exiting at different percentages of LLM layers used. }
\label{tab:early-exit}
    \vspace{-0.1in}
\end{table}

\paragraph{Compatibility.}
On the other hand, we can ideally build \modelname{} upon fine-tuned models to further improve its performance. However, the specific weights and architecture used to achieve SoTA results on the corresponding datasets are not publicly available, which are required for the development of \modelname{}.
Therefore, we use publicly accessible fine-tuned LLMs for this purpose. As shown in Table~\ref{tab:performance-finetuned}, \modelname{} consistently outperforms the corresponding baseline results across all datasets, demonstrating its effectiveness when applied to already fine-tuned and well-performing models. A rigorous statistical analysis confirming the significance of this improvement is provided in Appendix \ref{significance}. 

The scarcity of larger task-specific fine-tuned models here primarily due to the industry's shift of interest towards developing general-purpose LLMs, driven by the expensive costs of fine-tuning for specific tasks. This trend underscores the critical need for developing lightweight and scalable approaches, such as our proposed \modelname{}, to adapt LLMs to specialized tasks without incurring significant data and computational burdens.

\subsubsection{Inference Efficiency}
\label{inference-efficiency}

% \paragraph{Training Efficiency} Computational resources in the training stage can often pose significant hurdles in the adaptation of LLMs for specialized tasks. \modelname{} presents an efficient solution, both in terms of overall computational cost and resource utilization, making it a viable option even in resource-constrained environments. 

% Here we evaluate the theoretical computational cost within the training phase of \modelname{} with floating-point operations\footnote{See Appendix \ref{flo}.}\citep{kaplan2020scaling}. As shown in Figure \ref{fig:flo}, \modelname{} introduces only a marginal computational overhead on top of one forward pass of the LLM backbone -- which is the lower bound of adapting LLMs for text classification. 

% In addition, \modelname{} is a lightweight framework particularly in terms of trainable parameters. We estimate the numbers of trainable parameters of \modelname{} plugged in different baseline models and report them in Table \ref{}. Alongside the forward pass of LLMs, \modelname{} is not GPU-intensive, and translates directly into reduced energy consumption that leads to enhanced cost-effectiveness, which is a critical aspect in promoting sustainable AI practices.

% \rmv{
% This modest computational demand for \modelname{} training makes it easily accessible, especially in settings with constrained GPU resources for hosting LLM full fine-tuning, and translates directly into reduced energy consumption that leads to enhanced cost-effectiveness, which is a critical aspect in promoting sustainable AI practices\citep{kaplan2020scaling}.}

The cost for inference is crucial for real-world model deployment. We evaluate the effectiveness of \modelname{} with \textit{early-exit}, such that predictions are made before the entire forward pass finishes~\citep{pope2023efficiently, bae2023fast, chen2023ee}. In particular, only the internal neurons of part of the layers are sparsified and integrated to feed the classification head.
As shown in Table \ref{tab:early-exit}, \modelname{} enables early exiting in the top 60\% of layers, which ensures up to 99\% of the performance achievable when using all layers, significantly speeding up the inference process while maintaining high accuracy.
% The structural design of \modelname{} naturally supports this approach, as it can integrate neurons only from a fragment of the LLM backbone. 

% Following the discussion in \ref{agg}, we focus on the capability of the initial layers to capture task-specific features effectively. Table \ref{tab:early-exit} shows that the performance of \modelname{} fits this pattern, with diminishing returns as more layers of neurons are integrated. 

% This observation highlights the potential to exploit a trade-off between efficiency and performance during the inference process, which makes \modelname{} a practical solution in scenarios where the proportionate gain in efficiency is prioritized over the relatively subtle loss in performance.

\subsubsection{Interpretability}
% LLMs frequently come under scrutiny regarding their interpretability, as most current approaches for downstream tasks interact solely with their isolated terminal outputs. The limitation in understanding LLM's decision-making processes severely hinders their transparency-demanding application scenarios. Our \modelname{} framework addresses this concern from its structural design and post-hoc application process.

\begin{figure}[t]
    \centering
    \includegraphics[width=0.47\textwidth]
    {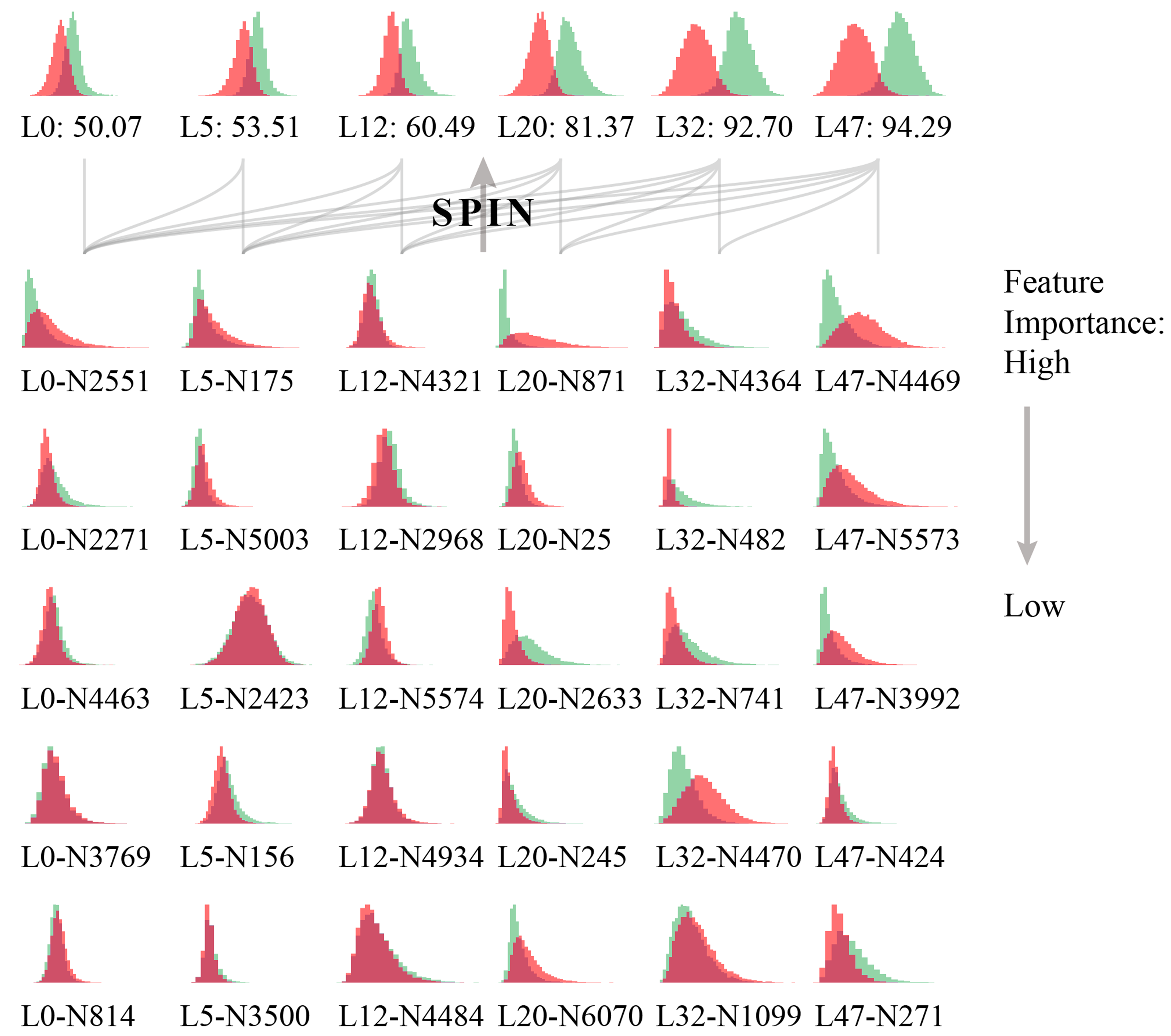}
    \vspace{0.1in}
    \caption{Activation probability distributions for individual salient neurons and integrated classifier at different layers of GPT2-XL. (Top) Distributions of SPIN with integrating neurons up to the specified layer, along with accuracy scores in text classification. (Bottom) Distributions of the most salient neurons according to their importance attributed by layer-wise neuron selection. Red regions indicate predictions for negative samples, and green regions for positive ones.} 
    \label{fig:intrinsic-interp}
    \vspace{-0.1in}
\end{figure}

\paragraph{Intrinsic Interpretability.} The exploration of the internal mechanisms of LLMs has laid a robust groundwork, from which we could investigate the rich interpretable data embedded within LLM neurons that is exploited by our \modelname{} framework. The sparsification process of \modelname{} can be viewed as a filtering of neurons, with both the training of linear regressors and the selection of salient neurons based on weight importance criteria inherently holding interpretability \citep{guyon2003introduction, Ceci2020New}. The integration process simply concatenates the selected salient neurons together, ensuring knowledge inherent in each individual salient neuron of LLM remains unchanged before being fed into the classification head. 

In Figure \ref{fig:intrinsic-interp}, we provide a visualized breakdown of how \modelname{}'s ability derives from the combination of neuron sparsification and integration. As is illustrated, each layer of the sparsified neurons collectively contributes their knowledge in differentiating positive and negative samples to the integrated classifiers thence and above, enhancing progressively the performance of \modelname{} at higher levels. This synergistic interaction empowers the overall decision-making and performance for the classification task.

\rmv{
Baseline models using terminal hidden states for text classification lack the interpretability because of their black-boxed nature. The implicit process of how they gather the capability to the classification head remains indistinct. On the contrary, \modelname{} is intrinsically interpretable with its linear probe-based sparsification and layer-by-layer integration process.
The inherent simplicity of linear probes emphasizes on task-relevant features rather than task-specific dynamics.

In the upper panel, we show the probability distribution of predict logits from \modelname{} at different layers.

As is illustrated, each layer of the sparsified neurons collectively contributes their knowledge in differentiating positive and negative samples to the integrated classifiers thence and above, enhancing progressively the performance of \modelname{} at higher levels.

}

\paragraph{Post-hoc Interpretability.} For a detailed post-hoc analysis of how \modelname{} interprets each component of the input text, we demonstrate that sentence-wise \modelname{} classifiers, once trained, can be adeptly extended to provide token-wise predictions by simply bypassing the initial pooling process and directly engaging with the representations of each individual token during application. Further analysis and discussion is continued in Appendix \ref{token-wise}. 

\begin{figure}[t]
    \centering
    \includegraphics[width=0.48\textwidth]{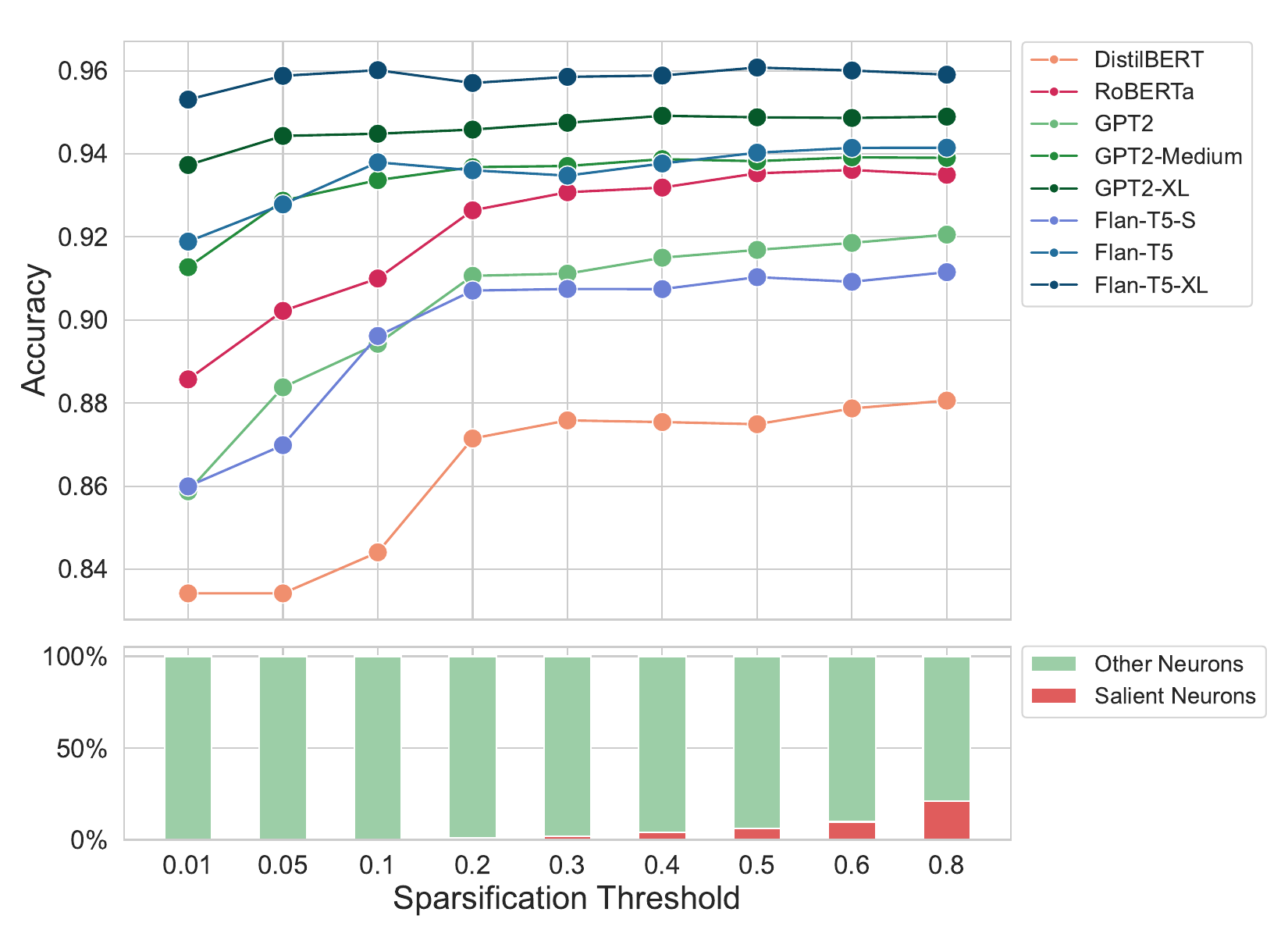}
    %\vspace{-0.2in}
    \caption{(Top) \modelname{} performance as a function of sparsification threshold $\eta$. 
 (Bottom) The percentage of selected salient neurons with each $\eta$. }
    \label{fig:hyper_sensitivity}
    \vspace{-0.1in}
\end{figure}

\subsubsection{Hyperparameter Sensitivity} 

As shown in Figure~\ref{fig:hyper_sensitivity}, with the increasing $\eta$, indicating that more neurons—yet less salient as sorted with decreasing weights—are being included, the performance gain gradually vanishes. The minimal accuracy gains beyond certain $\eta$ values suggest that the most salient features are already captured by a smaller, more focused set of neurons, and adding more neurons beyond this set contributes little to the overall performance. Especially, merely marginal performance improvement is observed when $\eta > 0.2$, demonstrating that \modelname{} can be well-performing with an easily found hyperparameter setting.
On the other hand, \modelname{} showcases a remarkable ability to select a highly compact subset of salient neurons as shown in the bottom half of Figure \ref{fig:hyper_sensitivity}. At a sparsification threshold of $\eta=0.4$, \modelname{} selects only about 3\% of neurons per layer. Despite this stringent selection, the performance remains highly competitive, underscoring the efficacy of our linear probing-based neuron selection method. Moreover, even at a relatively high threshold of $\eta=0.8$, the proportion of selected neurons does not exceed a quarter. This can be credited to the deployment of Lasso regressors, which effectively promotes feature sparsity.

\begin{table}[!t]
\small
\centering
\renewcommand\tabcolsep{5pt}
\renewcommand\arraystretch{0.9}
\begin{tabular}{lccccc}
\toprule
     & \multicolumn{5}{c}{\textbf{\#Neurons selected per layer}} \\
    \cmidrule{2-6}
     & 1 & 5 & 10 & 50 & 100  \\
\midrule 
    w/o SP  & 69.90 & 74.78 & 82.75 & 87.01 & 90.98 \\
    \modelname{} & 93.71 & 94.24 & 94.42 & 94.47 & 94.63\\
\bottomrule
\end{tabular}
\caption{Performance of \modelname{} and w/o SP, a variant of \modelname{} that uses random neuron selection without our designed sparsification strategy, across different numbers of neurons selected per layer.}
\label{tab:ablation_topk}
\end{table}

\begin{table}[!t]
\small
\centering
\renewcommand\tabcolsep{5pt}
\renewcommand\arraystretch{0.9}
\begin{tabular}{lccccc}
\toprule
    \textbf{\#Layer} & 8 & 16 & 24 & 32 & 48  \\
\midrule 
    w/o IN    & 87.17 & 89.84 & 92.72 & 94.12 & 92.41 \\
    \modelname{} & 88.34 & 92.09 & 94.70 & 94.91 & 94.92\\
\bottomrule
\end{tabular}
\caption{Performance of \modelname{} and w/o IN, a variant of \modelname{} that uses salient neurons from individual layers without our designed cross-layer integration strategy. We report the results of the best-performing individual layer across different early-exiting settings.}
\label{tab:ablation_probe}
\vspace{-0.1in}
\end{table}

\subsubsection{Ablation Study}
We conduct ablation studies with GPT2-XL on the IMDb dataset to verify the contribution of each component in \modelname{}. 
As shown in Table~\ref{tab:ablation_topk}, \modelname{} consistently outperforms the \modelname{} variant without our designed sparsification strategy across different numbers of neurons selected per layer (w/o IN), demonstrating the effectiveness of our linear probing-based neuron sparsification method.
In addition, we compare \modelname{} with its variant that without the cross-layer integration (w/o IN).
The results in Table~\ref{tab:ablation_probe} indicate the necessity of integrating cross-layer multi-grained features, as \modelname{} consistently outperforms w/o IN across various early-exiting settings.
These findings underscore the crucial roles that both sparsification and integration play within the \modelname{} framework.

\section{Related Work}
\label{related-work}

\paragraph{Deep Neural Networks in Text Classification.} The integration of Deep Neural Networks (DNNs) into text classification has significantly altered the methodological landscape of Natural Language Processing. An early study by \citet{kim2014convolutional} demonstrated the potential of Convolutional Neural Networks (CNNs) to capture semantic features from text. Research by \citet{conneau2017supervised} expanded the utility of DNNs through sentence embeddings and transfer learning, achieving advanced performance across multiple text classification benchmarks. Transformer-based models \citep{devlin2018bert} have systematically enhanced text classification by offering a versatile framework capable of understanding complex linguistic patterns. Studies \citep{yang2019xlnet, caselli2020hatebert} have since further pushed the performance of this approach. The success of ChatGPT marks the advent of a new generation of text classification using LLMs, with authors reporting mixed results \citep{susnjak2024applying, matter2024close, gilardi2023chatgpt}.

\paragraph{Interpretation of Language Models.} The rationale of SPIN primarily relates to the literature on interpreting internal representations in language models. Classic models like word2vec \citep{mikolov2013efficient} initially illustrated the linearly interpretable semantic features within the word embedding space. \rmv{The method of linear \emph{probes} to measure the concept understanding in neural networks is introduced by \citet{alain2016understanding}.} In transformer-based language models, the understanding of task-specific knowledge acquisition has been advanced by studies identifying internal neurons as \emph{experts} based on their activation patterns \citep{suau2020finding, durrani2020analyzing, burns2022discovering, gurnee2023finding}. Recent works \citep{bills2023language, templeton2024scaling} have also explored automated tools for evaluating the behaviors and interpretable features of individual neurons within modern LLMs. 

\paragraph{Internal Neurons for Text Classification.} Beyond interpretability, the prospect of leveraging internal representations for text classification has attracted attention in the literature. Studies as early as \citet{radford2017learning} have shown the potential of using individual neuron activations in Long Short-Term Memory (LSTM) models for sentiment classification. Relevant research including \citet{wang2022finding} empirically validated the potential of using the top-ranked task-specific neurons in RoBERTa~\cite{liu2019roberta} for classification, attaining performance competitive with fine-tuned models. \citet{gurnee2023language} demonstrates the effectiveness of leveraging internal representations in large models like LLaMA2~\cite{touvron2023llama} for spacial and temporal classification tasks. Inspired by these approaches targeting individual neurons, our work seeks more to design a general, efficient, plug-and-play framework for arbitrary types of transformer-based language models.

\paragraph{Dynamic Neural Network.} SPIN also shares design principles with previous works under the perspective of dynamic neural network \citep{han2021dynamic, xu2022survey}, such as adaptive parameter ensemble for CNN \citep{yang2019condconv}, Mixture-of-Experts (MoE) \citep{fedus2022switch}, and early exit strategies \citep{xin2020deebert, chen2023ee}. Specifically, like MoE, which exploits model sparsity by routing among multiple model components, SPIN leverages feature sparsification to enhance performance; while unlike MoE’s integrated routing process among FFN experts \citep{jiang2024mixtral}, SPIN operates as a decoupled, plug-and-play module at the granularity of FFN neurons. In terms of inference efficiency, SPIN aligns with early exit methods by allowing off-ramping at certain layers, as discussed in \ref{inference-efficiency} and Table \ref{tab:early-exit}.
%\textcolor{blue}{END OF PAGE LIMITATION PARTS}
\section{Conclusion}
In this paper, we introduce \modelname{}, a novel model-agnostic plug-and-play framework designed for text classification tasks. Our approach diverges from traditional paradigms that predominantly rely on the terminal hidden states of the final layer of LLMs by leveraging the untapped potential of internal neurons. Our proposed framework sparsifies neurons from intermediate layers guided by linear probing-based selection, and integrates cross-layer salient neurons to provide rich and multi-layered features for text classification. We conduct comprehensive experiments and analysis, demonstrating that \modelname{} remarkably improves accuracy, efficiency, and interpretability for text classification.

\newpage
\section*{Limitations} 
One of the primary limitations of the \modelname{} framework is its reliance on publicly available model architecture and weights, i.e., white-box LLMs. This requirement poses a challenge when working with some state-of-the-art (SoTA) models or proprietary fine-tuned models where the trained weights or the architecture are not publicly disclosed. As a result, while \modelname{} can theoretically be applied to various types of LLMs to potentially improve its performance, in practice, its deployment is limited to those models for which comprehensive access to internal mechanisms is publicly granted. Admittedly, the classification head can be improved using more sophisticated architectures, so as to further improve the text classification performance. However, we use the simple sigmoid activation on top of a linear transformation layer to better align with prior works and ensure interpretability.

\section*{Ethics Consideration}
\paragraph{Bias and fairness.} One of the primary ethical concerns in AI, particularly in natural language processing, is the potential for biased outcomes. LLMs, trained on large-scale internet data, can inadvertently learn and perpetuate biases present in the training data. \modelname{}, curating task-specific internal neurons of LLMs, could also be susceptible to these biases, and by construction \modelname{} is capable to discover potential biased components hidden inside LLMs. It is crucial to ensure that the model is not misused to amplify societal biases, particularly those related to race, gender, minority identity groups, or other sensitive attributes in different contexts and cultures. Continuous monitoring and mitigation strategies are necessary to address and reduce the impact of these biases, thereby promoting fairness and ethical responsibility in AI applications.

\paragraph{Dataset contents.} The datasets utilized in our research, namely IMDb, SST-2, AG News, and particularly EDOS (detecting online sexism), inherently contain a wide range of textual content, some of which might be sensitive or potentially harmful. The IMDb and SST-2 dataset reflect a broad spectrum of public opinions, and there is a possibility of encountering offensive or sensitive language, biased opinions, or controversial viewpoints. The EDOS dataset is explicitly designed to study online sexism, and as such, it contains examples of sexist remarks and content. Researchers must employ rigorous ethical standards, sensitivity, and transparency when working with these datasets. 

\section*{Acknowledgements}
We gratefully acknowledge the insightful comments and suggestions from our anonymous reviewers and area chair that helped us improve this manuscript. Part of this material is funded by grants from Natural Sciences and Engineering Research Council of Canada, Canada Foundation for Innovation, and Ontario Research Fund.

% \begin{itemize}
% %    \item Proxy classifiers are currently also Lasso regressors, which can be suboptimal.
%     \item \modelname{} only available on white-box LLMs; params of some fine tuned models or sota are not accessible
% %    \item \modelname{} vs LoRa (Adapter fine-tuning): Adapter working on frozen pretrained models could approach full-finetuning’s performance; \modelname{} need further results to compare
% %    \item \modelname{} vs LoRa (Adapter fine-tuning): for inference efficiency and interpretability, \modelname{} >>> LoRa
% %    \item \modelname{} vs LoRa (Adapter fine-tuning): in application, adapter contaminate terminal output hence lose the possibility of multi-tasking
% \end{itemize}

\rmv{
\subsection{Task Generalization}
\begin{itemize}
    \item can \modelname{} be used for other {'easy'} tasks than text generalization? text generator? more general tasks and benchmarks?
\end{itemize}

\subsection{ACL Checklist For Submission}
\begin{itemize}
    \item must have an individual section of LIMITATIONS. At the end of the paper, before the references, and it will not count toward the page limit.
    \item potential risks and ethical concerns?
    \item report the number of parameters in the models used, the total computational budget (e.g., GPU hours), and computing infrastructure used
\end{itemize}
}

\clearpage
\bibliography{custom}
\bibliographystyle{acl_natbib}

\clearpage
\appendix

\section{Additional Experiment Details}
\subsection{Dataset Description}
\label{appendix-dataset}

We select the following 3 datasets, as details summarized in Table \ref{tab:dataset}:
\begin{itemize}
    \item IMDb: The IMDb dataset \citep{maas2011learning} is one of the most popular sentiment classification datasets, curated for the binary classification task of positive and negative movie reviews. 
    \item SST-2: The SST-2 dataset for sentiment analysis, part of the General Language Understanding Evaluation (GLUE) benchmark \citep{wang2018glue}, provides a binary classification task based on the Stanford Sentiment Treebank. \rmv{The true test set labels of datasets within GLUE benchmark are not publicly accessible, and conventionally the fine-tuned models often report the performance on the validation set, hence we adopt the same approach by using the original dev set as the test set.}
    \item EDOS (SemEval-2023 Task 10): \citet{kirk2023semeval} collects dataset for facilitating exploratory experiments of Explainable Detection of Online Sexism (EDOS). The dataset contributes a hierarchical taxonomy of sexism content, in which we select Task A for our experiments, where systems are expected to predict whether a post is sexist or not.
\end{itemize}

\begin{table}[htbp]
\centering
\small
\begin{tabular}{rllr}
    \toprule
    Dataset & Subset & Label & \# Text \\
    \midrule
    IMDb & \textemdash & \verb+pos+, \verb+neg+ & 50,000\\
    \midrule
    GLUE & \verb+sst2+ & \verb+pos+, \verb+neg+ & 70,000\\
    \midrule
    \multirow{1}{*}{EDOS} & Task A & \verb+non_sexist+, \verb+sexist+ & 20,000 \\
    \bottomrule
\end{tabular}
\caption{All datasets and features used}
\label{tab:dataset}
    \vspace{-0.1in}
\end{table}

\subsection{Hyperparameter Settings}
\label{hyperparams}
Here we provide the hyperparameter search space across our experiments with \modelname{} in Table \ref{tab:hyper_space}, and the corresponding hyperparameter setting in Table \ref{tab:best_hyper}, to facilitate a better reproducibility of our reported results. 

\begin{table}[!t]
    \centering
    \small
    \begin{tabularx}{.48\textwidth}{lX}
    \toprule
        Representation & hidden state, activation \\ 
        Pooling choice  & first, last, max, avg\\
        $\lambda$       & \{0.01, 0.1, 1, 5, 10\}\\
        $\eta$          & \{0.001, 0.005, 0.01, 0.05, 0.1, 0.2, 0.3, 0.4, 0.5, 0.6, 0.8, 1\} \\
        %Iterations      & \{0, 1, 2, ..., 64\}\\
    \bottomrule
    \end{tabularx}
    \caption{Hyperparameter search space of \modelname{}} %\textcolor{orange}{remove iteration}}
    \label{tab:hyper_space}
\end{table}

\begin{table*}[!t]
    \centering
    \small
    \begin{tabular}{l|cccc|cccc|cccc}
    \toprule
         & \multicolumn{4}{c|}{\textbf{IMDb}} & \multicolumn{4}{c|}{\textbf{SST-2}} & \multicolumn{4}{c}{\textbf{EDOS}}  \\
        \cmidrule(l{3pt}r{3pt}){2-5} \cmidrule(l{3pt}r{3pt}){6-9} \cmidrule(l{3pt}){10-13}
        \textbf{LLM} & rep. & pool & $\lambda$ & $\eta$ & rep. & pool. & $\lambda$ & $\eta$ & rep. & pool. & $\lambda$ & $\eta$ \\
    \midrule
        DistilBERT  & act & avg & 5   & 1.0 & act & avg & 10  & 1.0 & hs  & max & 5   & 0.8 \\
        RoBERTa     & act & avg & 10  & 0.5 & act & avg & 10  & 0.6 & hs  & max & 5   & 0.5 \\
        GPT2        & act & avg & 10  & 0.6 & act & avg & 5   & 0.8 & act & avg & 5   & 0.5 \\
        GPT2-M      & act & avg & 1   & 0.6 & act & max & 1   & 0.6 & act & avg & 1   & 0.4 \\
        GPT2-XL     & act & avg & 0.2 & 0.4 & act & avg & 0.5 & 0.3 & act & avg & 0.5 & 0.2 \\
        Flan-T5-S   & act & avg & 1   & 0.8 & act & avg & 1   & 0.6 & act & max & 0.5 & 0.6 \\
        Flan-T5     & act & avg & 0.2 & 0.6 & act & avg & 0.5 & 0.6 & act & avg & 0.5 & 0.4 \\
        Flan-T5-XL  & act & avg & 0.2 & 0.3 & act & avg & 0.2 & 0.4 & act & avg & 0.2 & 0.2 \\
%        GPT2-L      & act & max &   & 0.8 &\\
%        Flan-T5-L   & act & avg &   & 0.8 &\\
        \bottomrule
    \end{tabular}
    \caption{Hyperparameter settings of \modelname{} over frozen pretrained LLMs}
    \label{tab:best_hyper}
    \vspace{-0.1in}
\end{table*}

\section{Additional Experiment Analysis}

\subsection{Trainable Parameters}
\label{param}
We compare the number of trainable parameters between baseline LLMs and our \modelname{} framework. For baseline LLMs we refer to the model size reported by huggingface safetensors. For \modelname{}, we derive our estimation by calculating the total parameters of linear probes and classification head used at different stages of \modelname{}:
\begin{align}
    N_\text{param SPIN}\approx LD + \frac{1}{2}L^2(\rho_\eta D)
\end{align}
where $L$ refers to the number of layers of the LLM, $D\in \{D_{\text{hs}},D_{\text{act}}\}$ the actual dimension of internal representation used, and $\rho_\eta$ the ratio of neurons sparsified by the salient neuron selection. Here the first half refers to the layer-wise salient neuron selection process, where we trained $L$ times individual linear probes, each of which with $D$ trainable parameters. The second half takes a typical workflow of aggregation across all layers, with $\frac{L(L+1)}{2}$ times training of classification heads with sparsified neurons.

\begin{figure}[!t]
    \centering
    \includegraphics[width=0.48\textwidth]
    {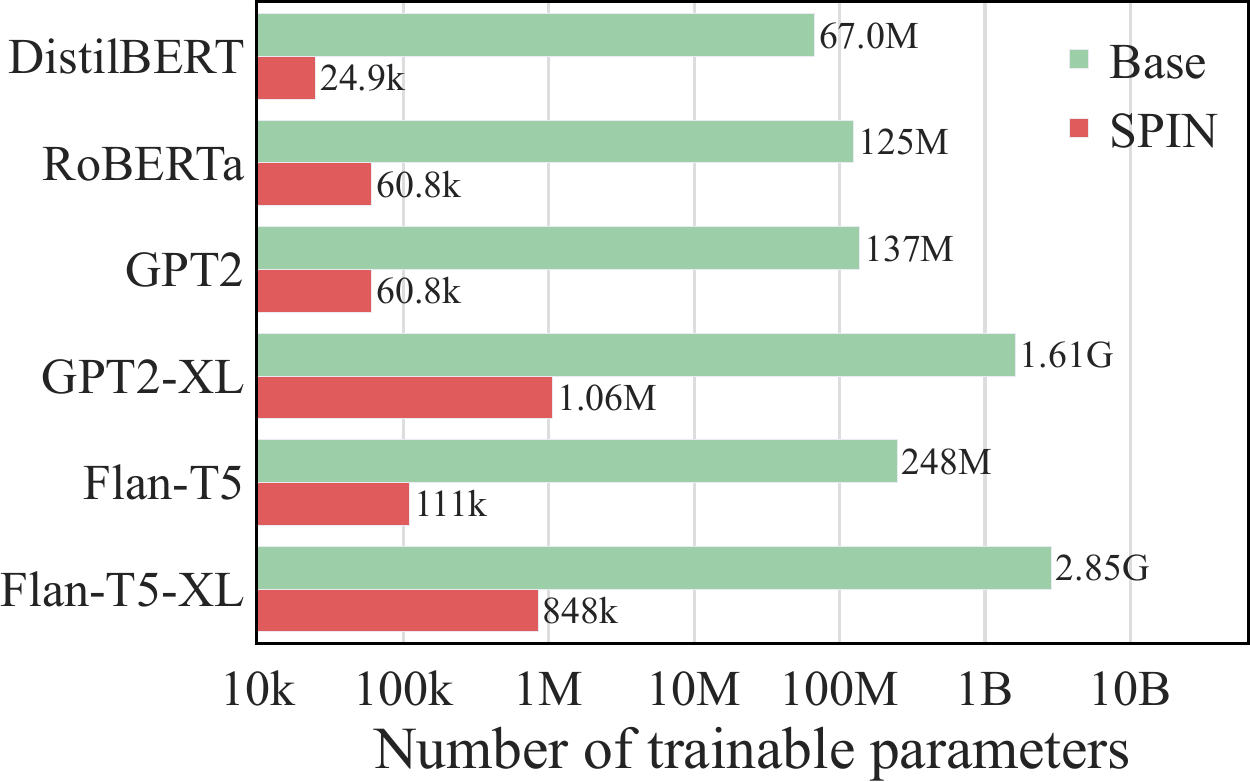}
    \caption{The number of trainable parameters for baseline LLM backbone and \modelname{}.}
    \label{fig:N_params}
    \vspace{-0.1in}
\end{figure}

The estimated results are provided in Figure \ref{fig:N_params}, showing that the training process of \modelname{} yields at least three orders of magnitude fewer parameters than the LLM it works on.

\subsection{Floating Point Arithmetic}
\label{flo}
In Figure \ref{fig:flo} of training efficiency, for the total floating point operations required in running LLM forward pass, we refer to the empirical estimations by \citet{kaplan2020scaling}, from which we have 
\begin{align}
    C_{\text{LLM forward}} \approx(2N_{\text{param}}+2LD_{\text{hs}}N_{\text{token}})\cdot N_{\text{s}}
\end{align}
where $N_{\text{param}}$ represents the total amount of parameters with LLM, which we refer to the size reported by huggingface safetensors for each model; $L$ the number of layers, $D_{\text{hs}}$ the dimension of hidden states, $N_{\text{token}}$ the maximum number of tokens for model input, and $N_{\text{s}}$ the number of sentences within the training and validation set.

For the floating point operations required in our \modelname{} framework, we derive from the training cost of a single linear regression:
\begin{align}
    C_{\text{LR train}} &\approx I \cdot (2D\cdot N_{\text{s}} + N_{\text{s}}) 
\end{align}
where $I$ refers to the maximum number of iterations for the training of each linear regression, and $D\in \{D_{\text{hs}},D_{\text{act}}\}$ the actual dimension of internal representation used. The $2D\cdot N_{\text{s}}$ part is for the multiplication and additions required in gradient computation, and the $N_{\text{s}}$ part is for applying logistic functions on output, which becomes marginal comparing with the dimension of input features. The parameter update process is generally negligible in comparison to the gradient computation. 

By incorporating $L$ times $C_{\text{LR train}}$ for salient neuron selection process and $\frac{L(L+1)}{2}$ times $C_{\text{LR train}}$ for a typical workflow of integration across all layers, we get an approximation of
\begin{align}
    C_{\text{\modelname{}}} &\approx I \cdot 2LD + L^2 (\rho_\eta D))\cdot N_{\text{s}}
\end{align}
in which $\rho_\eta$ the ratio of neurons sparsified by the salient neuron selection.

The floating point operation cost reported in Figure \ref{fig:flo} takes $I=64$, $D=D_{\text{act}}$, $\rho_\eta=0.1$, $N_\text{s}=25000$ for IMDb dataset, and other values according to the corresponding model settings.

\section{Additional Results and Discussion}

\subsection{Parameter-Efficient Fine-Tuning}
\label{PEFT}
In this section, we discuss briefly on the compatability and comparability of SPIN with Parameter-Efficient Fine-Tuning (PEFT) techniques. 

\paragraph{Compatibility.} 
SPIN is by construction compatible with LLMs integrated with PEFT methods, since SPIN relies solely on the internal representations within the model, regardless of whether and how they are fine-tuned. According to \citet{he2021towards}, mainstream PEFT techniques typically introduce additional structures into LLMs either sequentially between transformer block components (as with adapters \citep{houlsby2019parameter}) or parallelly alongside transformer block components (as in LoRA \citep{hu2021lora}, prefix tuning \citep{li2021prefix}, etc.), none of which hinders SPIN’s process to acquire internal representations from FFN activations and hidden states. Depending on the extent of modifications to the LLM’s internal mechanisms, SPIN requires no or minimal adjustments to continue functioning as a plug-and-play module over PEFT-modified LLMs. Here we show results from an experiment using a LoRA-finetuned DistilBERT model available on HuggingFace over the IMDb dataset as an example. The performance results presented in Table \ref{tab:lora} demonstrate the effectiveness of SPIN on PEFTed LLMs.

\begin{table}[t]
\centering
\small
\begin{tabular}{lccc}
    \toprule
    Models     & Base & SPIN & \%impr. \\
    \midrule
    DistilBERT Pretrained & 86.95 & \textbf{89.78} & +3.25 \\
    \textit{DistilBERT w/ LoRA SFT} & 87.71 & \textbf{90.58} & +3.27 \\
    DistilBERT w/ Full SFT  & 92.80 & \textbf{92.88} & +0.09 \\
    \bottomrule
\end{tabular}
\caption{Performance of SPIN and baseline methods (Base) over pretrained, LoRA supervised fine-tuned (LoRA SFT), and full fine-tuned (Full SFT) DistilBERT for IMDb dataset.}
\label{tab:lora}
\vspace{-0.1in}
\end{table}

\paragraph{Comparability.}
A key distinction between SPIN and PEFT methods is that SPIN is completely decoupled from LLMs. Here we present a series of comparisons across several dimensions of concern.
\begin{itemize}
    \item \textbf{Performance.} Both SPIN and PEFT achieve performance approaching fully fine-tuned models when working over pretrained LLMs.
    \item \textbf{Parameter Efficiency.} Both methods share a similar scale of trainable parameters. PEFT requires hosting the entire LLM during the whole training process for both the forward and backward passes. In contrast, SPIN's training process, apart from recording internal representations, can occur in highly computationally constrained environments, requiring only a few classifiers to be updated.
    \item \textbf{Training Time Efficiency.} PEFT components are by design highly coupled with the LLM. When applied to pretrained LLMs, though only the set of added parameters is updated, gradient computation must still cascade through the entire network, resulting actually much more time than floating-point operations in Figure \ref{fig:flo} suggest \citep{sun2023comparative}. In contrast, SPIN trains lightweight linear probes independently from the LLM over the recorded internal representations, hence the floating-point operations can faithfully reflect the computation time required. Additionally, SPIN's training can be more easily accelerated in parallel compared to PEFT methods.
    \item \textbf{Inference Efficiency.} SPIN naturally supports a range of dynamic neural network methods like early-exit by introducing no or minimal adjustments, whereas PEFT does not.
    \item \textbf{Interpretability.} Neither PEFT nor full fine-tuning offers interpretability comparable to that of SPIN.
\end{itemize}

\rmv{shows that LoRA with 0.26\% trainable parameters of LLaMA-7B requires approximately 23\% of the training time per epoch compared to full fine-tuning.}

\subsection{Multiclass Classification}
\label{multiclass}

\begin{table}[t]
\centering
\small
\begin{tabular}{lccc}
    \toprule
    Models & Base & SPIN & \%impr. \\
    \midrule
    DistilBERT & 91.18 & \textbf{92.21} & +1.13 \\
    GPT2       & 90.64 & \textbf{92.19} & +1.71 \\
    Flan-T5-S  & 85.68 & \textbf{91.89} & +7.24 \\
    \bottomrule
\end{tabular}
\caption{Performance (accuracy) of SPIN and baseline method (Base) over LLMs for AG News dataset.}
\label{tab:ag-news}
\vspace{-0.1in}
\end{table}

Our experiments were all implemented on classification tasks with binary features. Classification over multiclass labels (categories) can easily be transformed to multiple binary classifications by representing labels as one-hot encoded binary features and independently training one binary classifier for each feature \citep{read2011classifier}. Another approach is by adapting our classification heads into algorithms that naturally permit usage of more than two classes. Here we test SPIN’s performance on AG's news topic classification dataset \citep{zhang2015character} and train Multinomial Logistic Regressor as salient neuron selector and classification head, as reported in Table \ref{tab:ag-news}.

\subsection{Scalability}
\label{scalability}

As mainstream LLMs continue to increase in scale, in this part we showcase the scalability of SPIN on larger models LLaMA2-7B and LLaMA2-13B \citep{touvron2023llama}. Results shown in Table \ref{tab:llama2} suggest that SPIN consistently outperforms the model outputs by exploiting the ability of internal neurons even from large models with 10B-level parameters. This is particularly significant, considering that fine-tuning modern LLMs on simpler tasks as sentence classification and with limited training examples becomes increasingly uneconomical and impractical as they continue scaling-up. SPIN offers a viable solution, an efficient and effective means to boost performance without extensive additional resources and processes.

\begin{table}[t]
\centering
\small
\begin{tabular}{lccc}
    \toprule
    Models     & Base & SPIN & \%impr. \\
    \midrule
    LLaMA2-7B  & 94.04 & \textbf{95.76} & +1.83 \\
    LLaMA2-13B & 94.55 & \textbf{96.06} & +1.60 \\
    \bottomrule
\end{tabular}
\caption{Performance of SPIN and baseline methods (Base) over pretrained LLaMA2 models for IMDb dataset.}
\label{tab:llama2}
\end{table}

\subsection{Token-wise Classification}
\label{token-wise}

\begin{figure}[t]
    \centering
    \subfloat[Base]{
        \includegraphics[width=0.46\textwidth, clip]{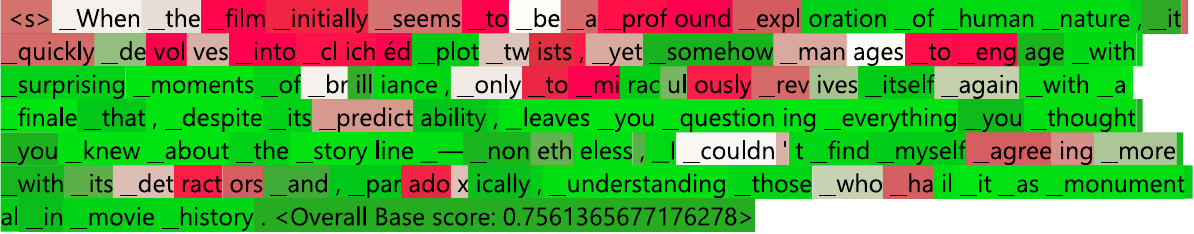}
    }\\
    \subfloat[SPIN]{
        \includegraphics[width=0.46\textwidth, clip]{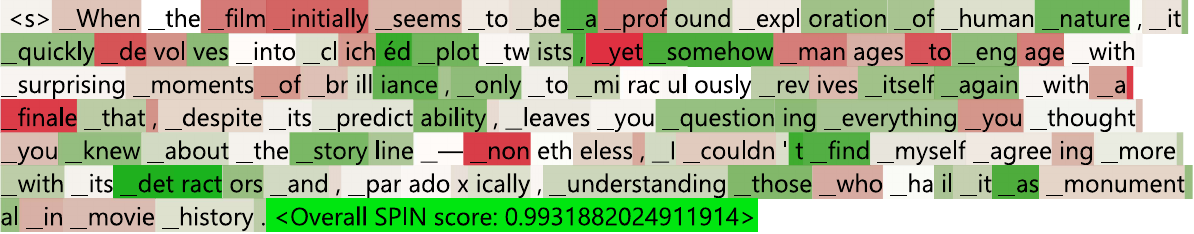}
    }
    \caption{Transferred token-wise breakdown of classification results, with (a) baseline and (b) SPIN originally trained on sentence-wise classification examples from IMDb using GPT2-XL neurons and max pooling. Red regions indicate predictions for negative tokens, and green regions for positive ones.}
    \label{fig:post-hoc-interp-base-spin}
\vspace{-0.1in}
\end{figure}

We attribute SPIN's adaptability of transferring from sentence-wise to token-wise classification to the introduction of max and average pooling strategies during its training phase. These pooling processes enables \modelname{} initially trained on broader textual scopes to recognize similar patterns at the token level. 

For encoder-based BERT variants, \modelname{} facilitates a previously unavailable transfer capability. Conventional baseline methods structurally rely on the first or special token for information gathering, making direct transfer to token-level tasks infeasible. These methods often require repetitive sentence inputs with sliding context windows for transferring, which leads to significantly higher computational costs and time. In contrast, \modelname{} overcomes this limitation by effectively utilizing the internal representations from all tokens at the beginning in training sentence-level classification.

For decoder-based variants, \modelname{} trained on pooled records exhibits similar ability to conventional methods trained on the last token's output. This allows for direct transfer from sentence-level to token-level tasks, as demonstrated in Figure \ref{fig:post-hoc-interp-base-spin}. The extended \modelname{} prediction results align well with the fine-grained, \textit{cumulative} sentiment expressed at each token. 

It is important to note that the use of causal attention in decoder models means that the activations and hidden states of each token represent information from \textit{all preceding positions}, unlike the bidirectional understanding in encoder-based models. This characteristic enables decoder models to aggregate context in a sequentially cumulative manner. For instance, in Figure \ref{fig:post-hoc-interp-base-spin}, during the token positions of ``couldn't find myself agreeing more with'', both the baseline method and SPIN accurately capture multiple times of sentiment flipping to the opposite, as expected in a cumulative sentiment sequence. At word positions like ``yet somehow'' ``despite'' and ``nonetheless'', the result of SPIN shows much clearer patterns than the baseline method. This alignment facilitates better for a transparent post-hoc breakdown of how each token contributes to the overall sentence-level classification, showcasing the robustness of our framework.

\begin{table}[t]
\centering
\small
\begin{tabular}{lccl}
    \toprule
    Models     & Base & SPIN & $p$-value \\
    \midrule
    DistilBERT & 92.80 & \textbf{92.86}$\pm 0.0371$ & 0.0169 \\
    RoBERTa    & 94.67 & \textbf{95.62}$\pm 0.0425$ & $8\times 10^{-7}$\\
    GPT2       & 94.06 & \textbf{94.47}$\pm 0.1224$ & 0.0014 \\
    \bottomrule
\end{tabular}
\caption{Performance of SPIN ($k$=5 fold cross-validation) and baseline methods (Base) over fine-tuned LLMs for IMDb dataset, with 95\% confidence interval ($z$=1.96) and corresponding $p$-value result from $t$-tests.}
\label{tab:cross-validation}
\vspace{-0.1in}
\end{table}

By bridging the gap across different transformer-based model architectures and offering previously obscured insights, \modelname{} holds significant potential as a practical tool for visualizing complex sentences or documents with classification results at varying levels of granularity. This capability lays the foundation for more transparent and accountable systems for end-users.

\subsection{Statistical Significance}
\label{significance}

To ensure consistency, our aforementioned experiments were conducted using the same train/validation/test splits as established benchmarks. In order to further validate the effectiveness of our proposed SPIN framework, especially regarding the relatively marginal gains over fine-tuned models as shown in Table \ref{tab:performance-finetuned}, we hereby perform a rigorous statistical significance evaluation using $k$-fold cross-validation. As detailed in Table \ref{tab:cross-validation}, the constant performance of the baseline model (Base) from the original dataset split is used as the mean value for the null hypothesis. We then apply the one-sample Student's $t$-test on the 5-fold cross-validation results of SPIN, in testing the hypothesis that SPIN can further improve the performance of an already fine-tuned model. Each performance result yields a $p$-value less than 0.05, indicating that the improvements achieved by SPIN are statistically significant across the three models tested.

\subsection{What-Which-Where Plots}
\label{www}

For given LLMs, we train and evaluate the classifier on integrated features with each combination of pooling function $\text{Pooling}(x_l)$, the sparsification threshold $\eta$, and the number of layers integrated $L$. We refer to following figures as \emph{what-which-where} plots for better visualizing and interpreting the performance of \modelname{} on GPT2-XL over the IMDb dataset, along with information of \emph{what} sparsification threshold is used, \emph{which} pooling function is selected, and \emph{where} in depth the layer integration (i.e., early exiting) takes place in LLMs. Respectively, the pooling choice is indicated with different marker shapes, the magnitude of $\eta$ by the marker sizes, and the number of integrated layers is shown with different colors as described in the color bar above. The horizontal axes represent the number of salient neurons integrated. Additionally, in the lower panels, we present how the number of salient features evolves with different level of layer integration, and on the left sides of each figure are the performance of \modelname{} grouped by different pooling functions used.

\rmv{
\begin{figure*}[htbp]
    \centering
    \subfloat[Pretrained model]{
        \includegraphics[width=\textwidth]{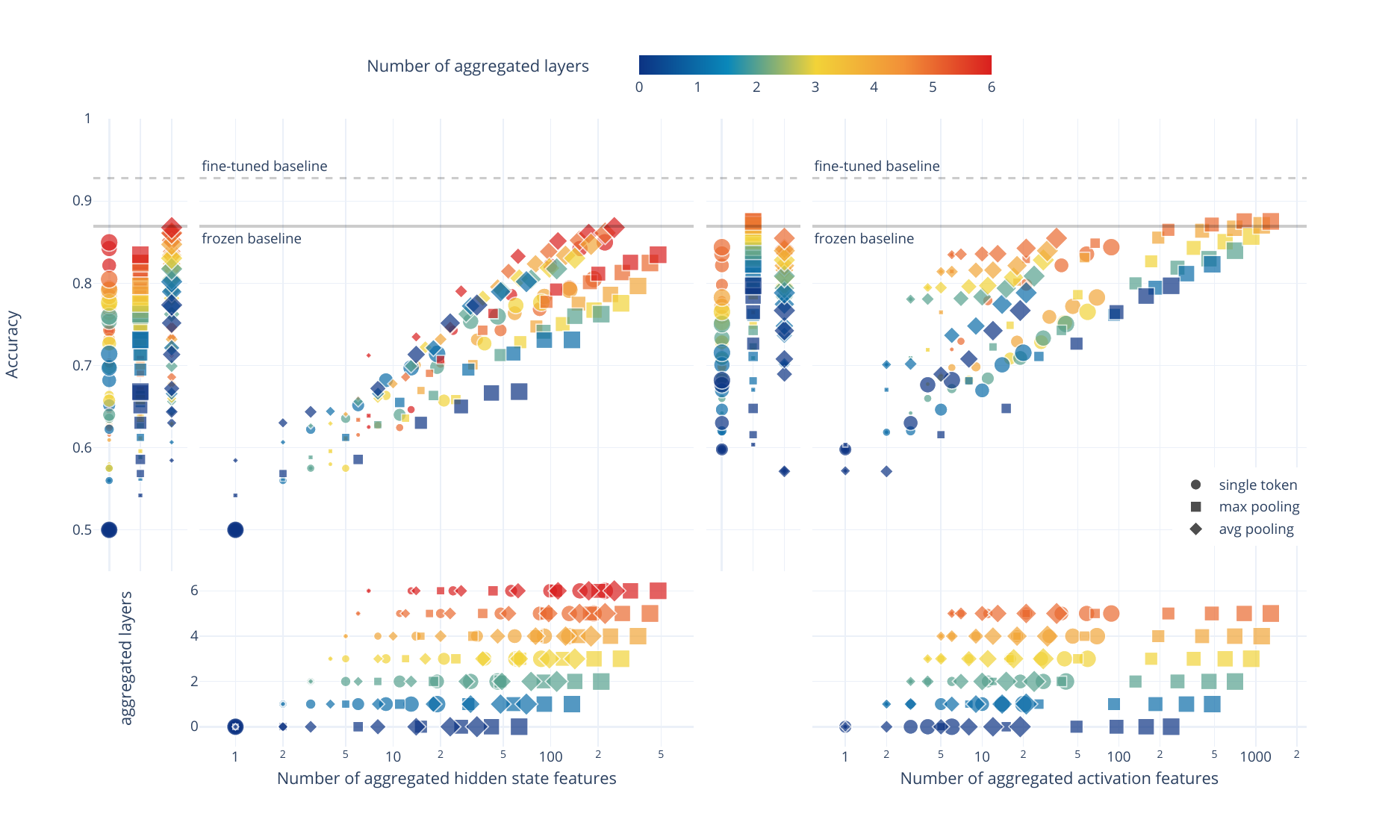}}
    \\
    \subfloat[Fine-tuned model]{
        \includegraphics[width=\textwidth]{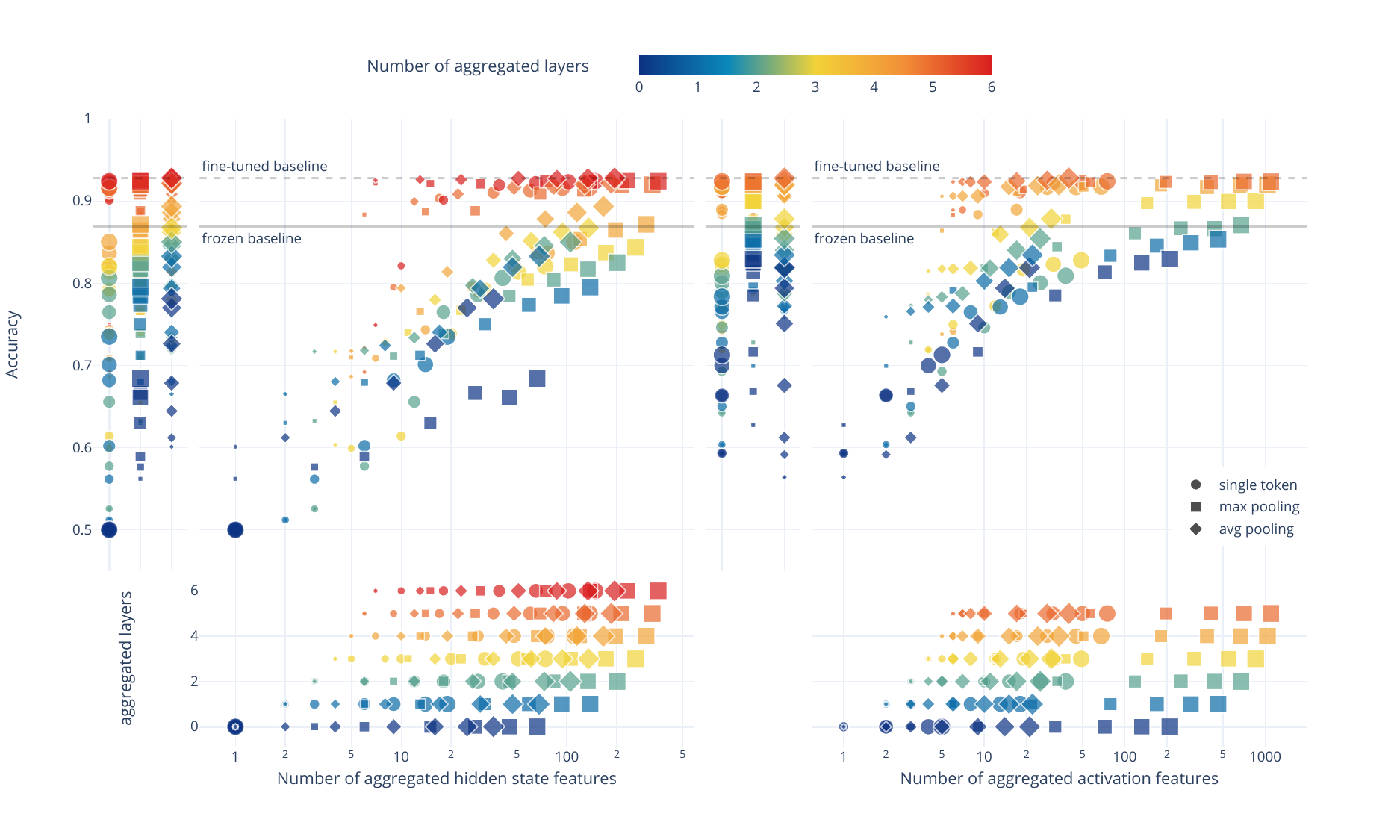}}
    \caption{What-which-where plots for the performance of \modelname{} on pretrained and fine-tuned DistilBERT models over IMDb dataset. The left subfigures are results with hidden states, and the right subfigures are with activations.}
\end{figure*}

\begin{figure*}[htbp]
    \centering
    \subfloat[Pretrained model]{
        \includegraphics[width=\textwidth]{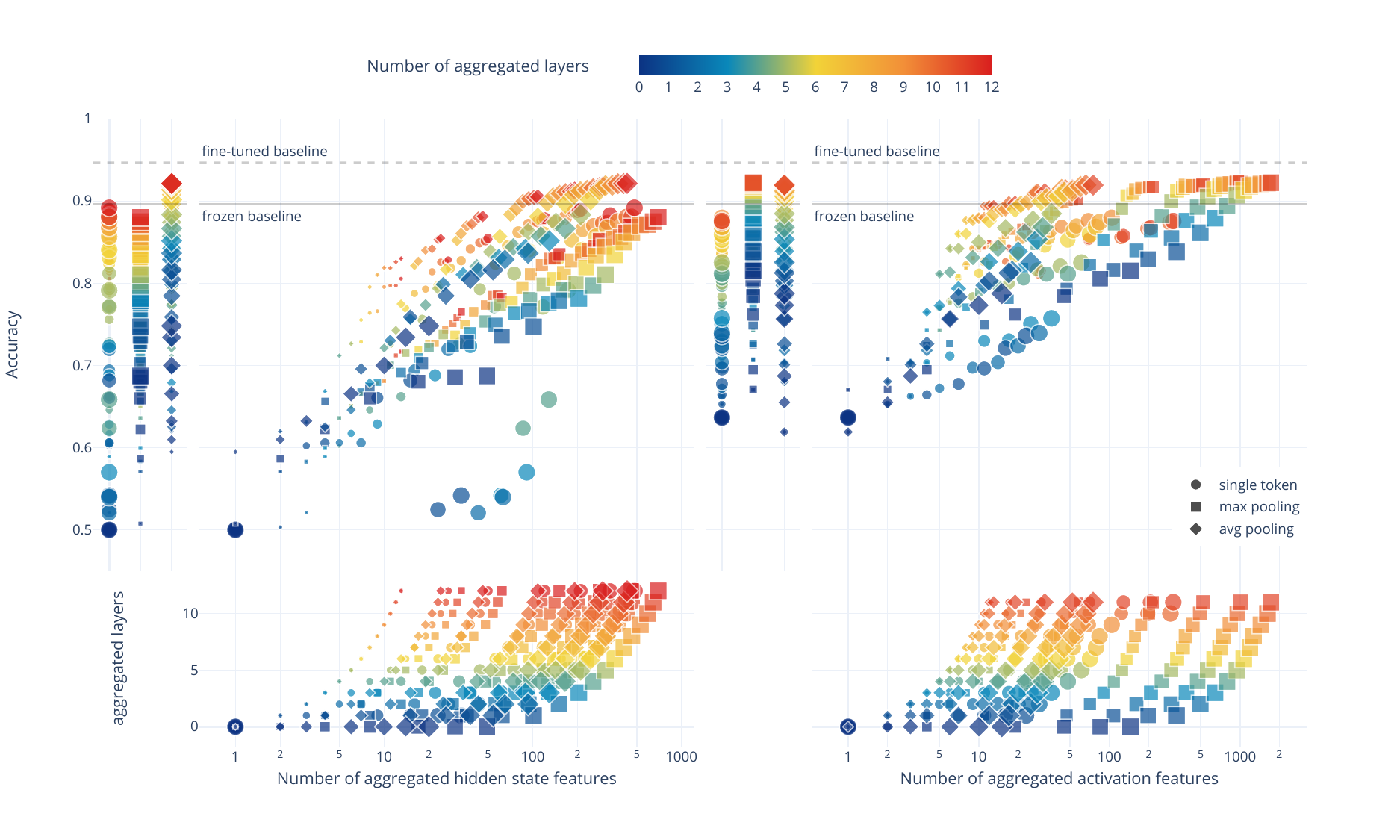}}
    \\
    \subfloat[Fine-tuned model]{
        \includegraphics[width=\textwidth]{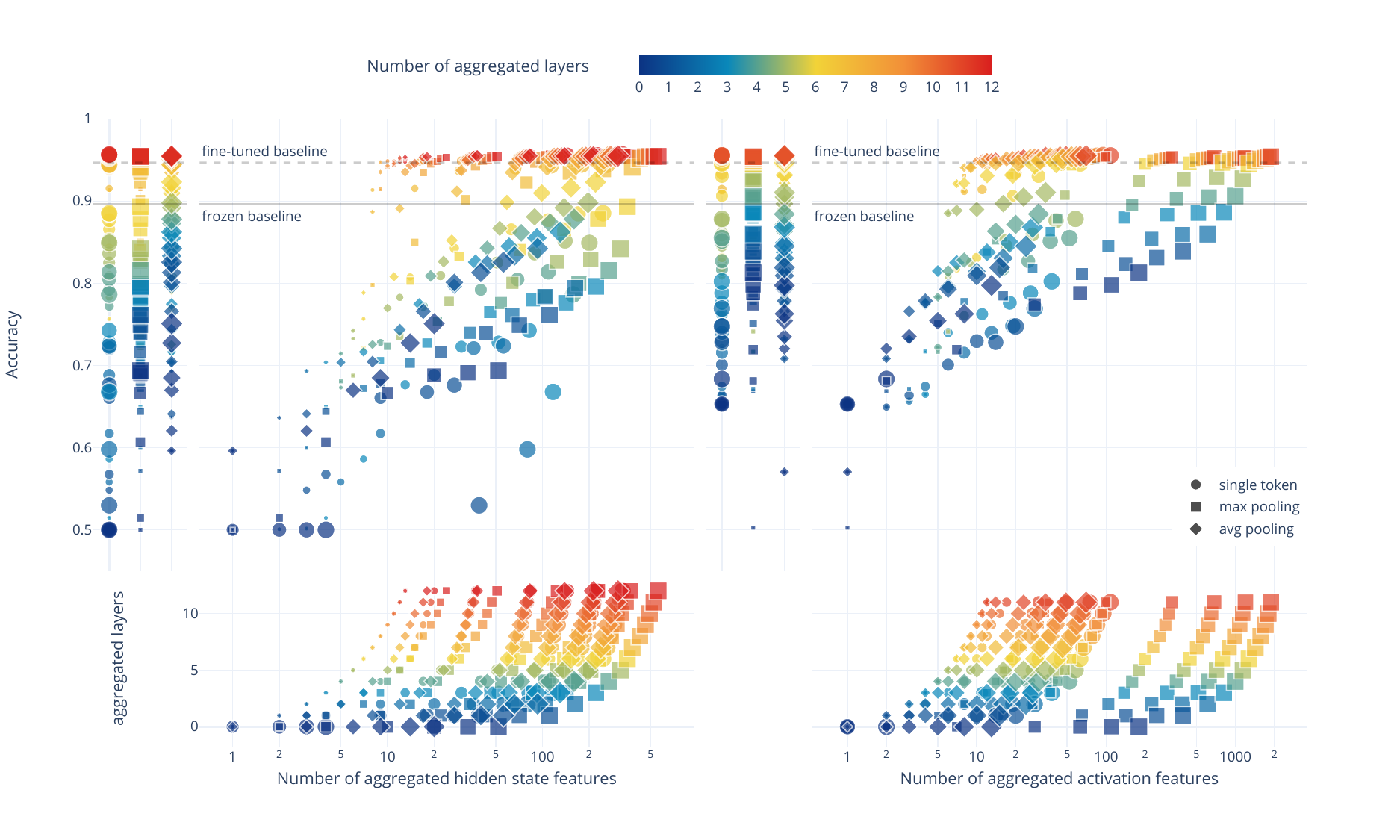}}
    \caption{What-which-where plots for the performance of \modelname{} on pretrained and fine-tuned RoBERTa models over IMDb dataset. The left subfigures are results with hidden states, and the right subfigures are with activations.}
\end{figure*}

\begin{figure*}[htbp]
    \centering
    \subfloat[Pretrained model]{
        \includegraphics[width=\textwidth]{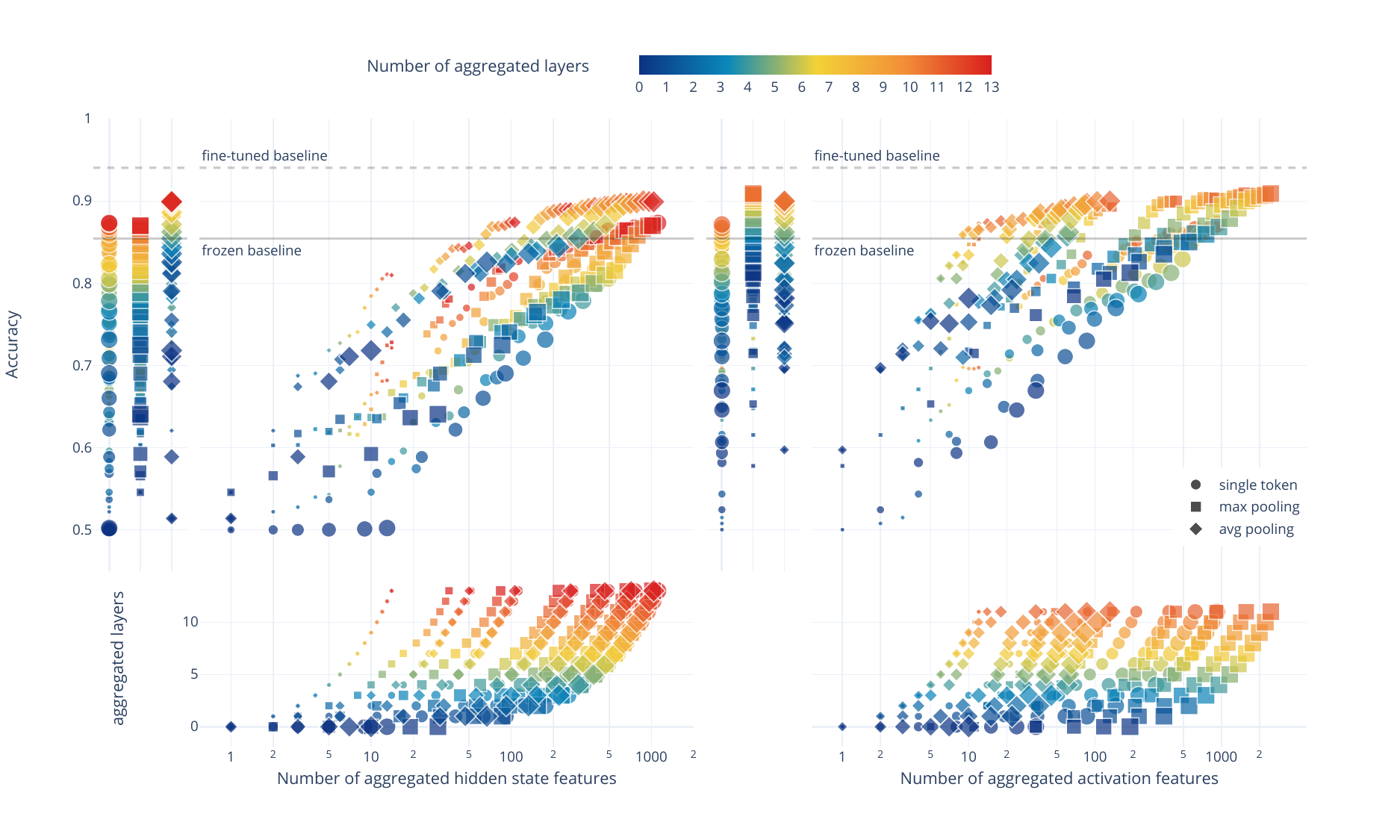}}
    \\
    \subfloat[Fine-tuned model]{
        \includegraphics[width=\textwidth]{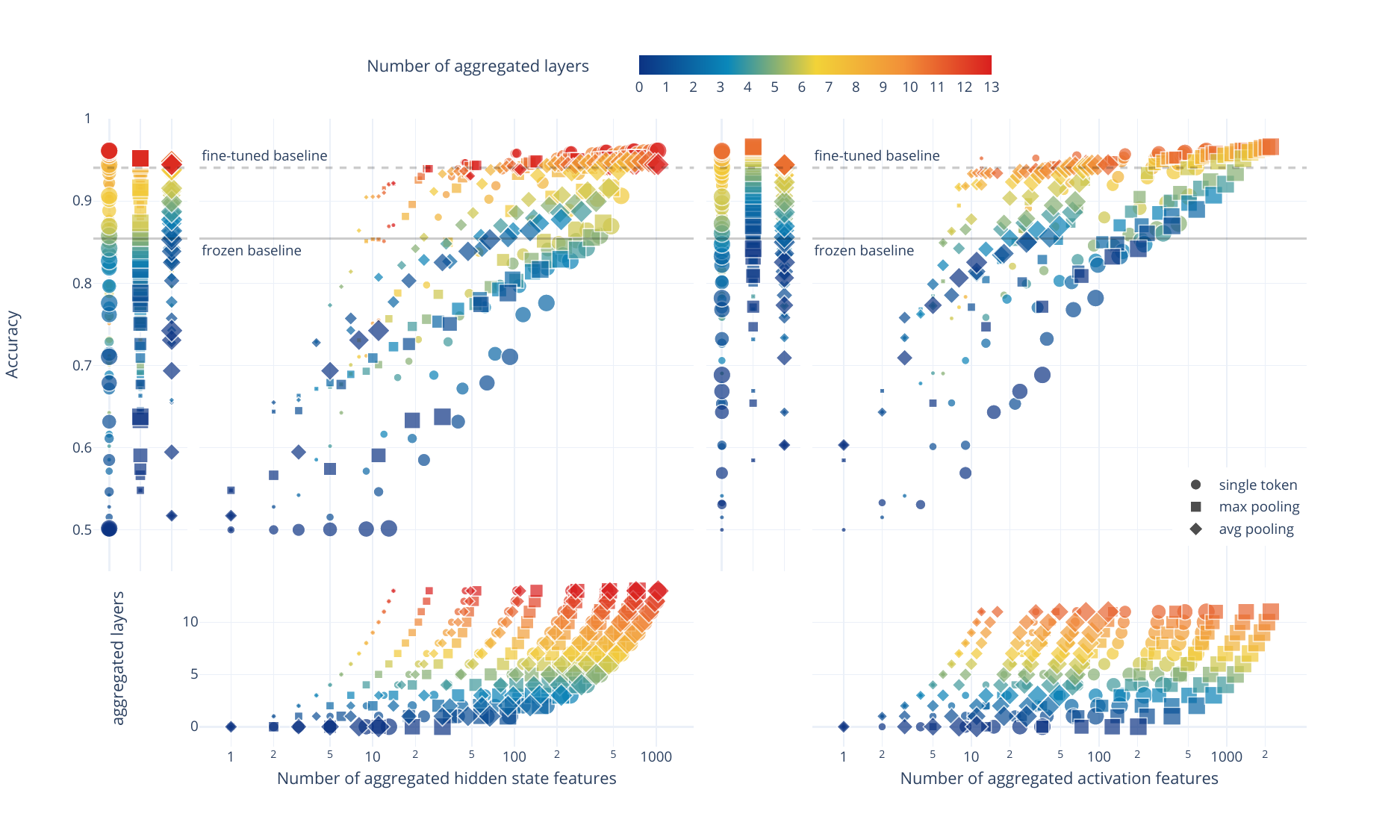}}
    \caption{What-which-where plots for the performance of \modelname{} on pretrained and fine-tuned GPT2 models over IMDb dataset. The left subfigures are results with hidden states, and the right subfigures are with activations.}
\end{figure*}

\begin{figure*}[htbp]
    \centering
    \subfloat[Pretrained model]{
        \includegraphics[width=\textwidth]{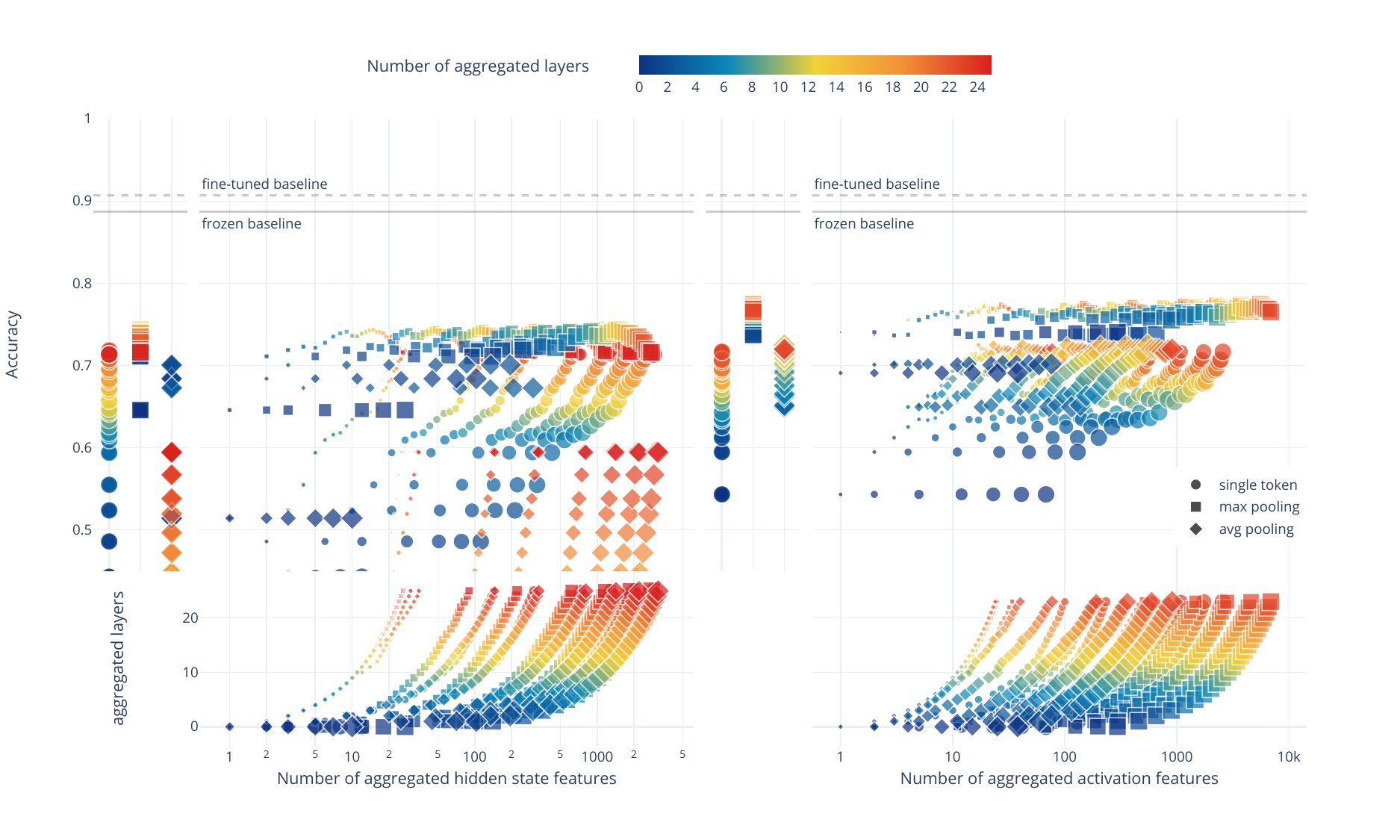}}
    \\
    \subfloat[Fine-tuned model]{
        \includegraphics[width=\textwidth]{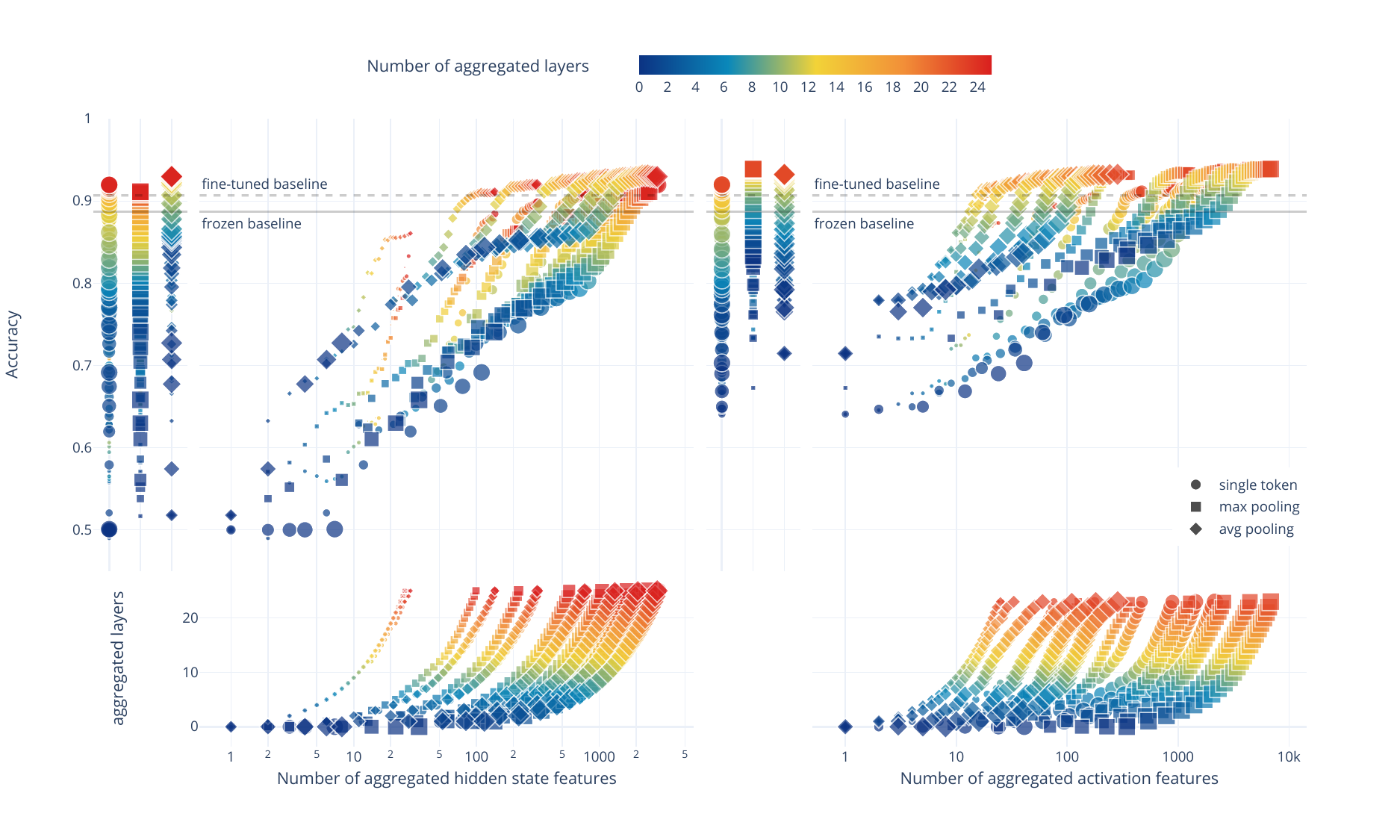}}
    \caption{What-which-where plots for the performance of \modelname{} on pretrained and fine-tuned GPT2-M models over IMDb dataset. The left subfigures are results with hidden states, and the right subfigures are with activations.}
\end{figure*}

\begin{figure*}[htbp]
    \centering
    \subfloat[Pretrained model]{
        \includegraphics[width=\textwidth]{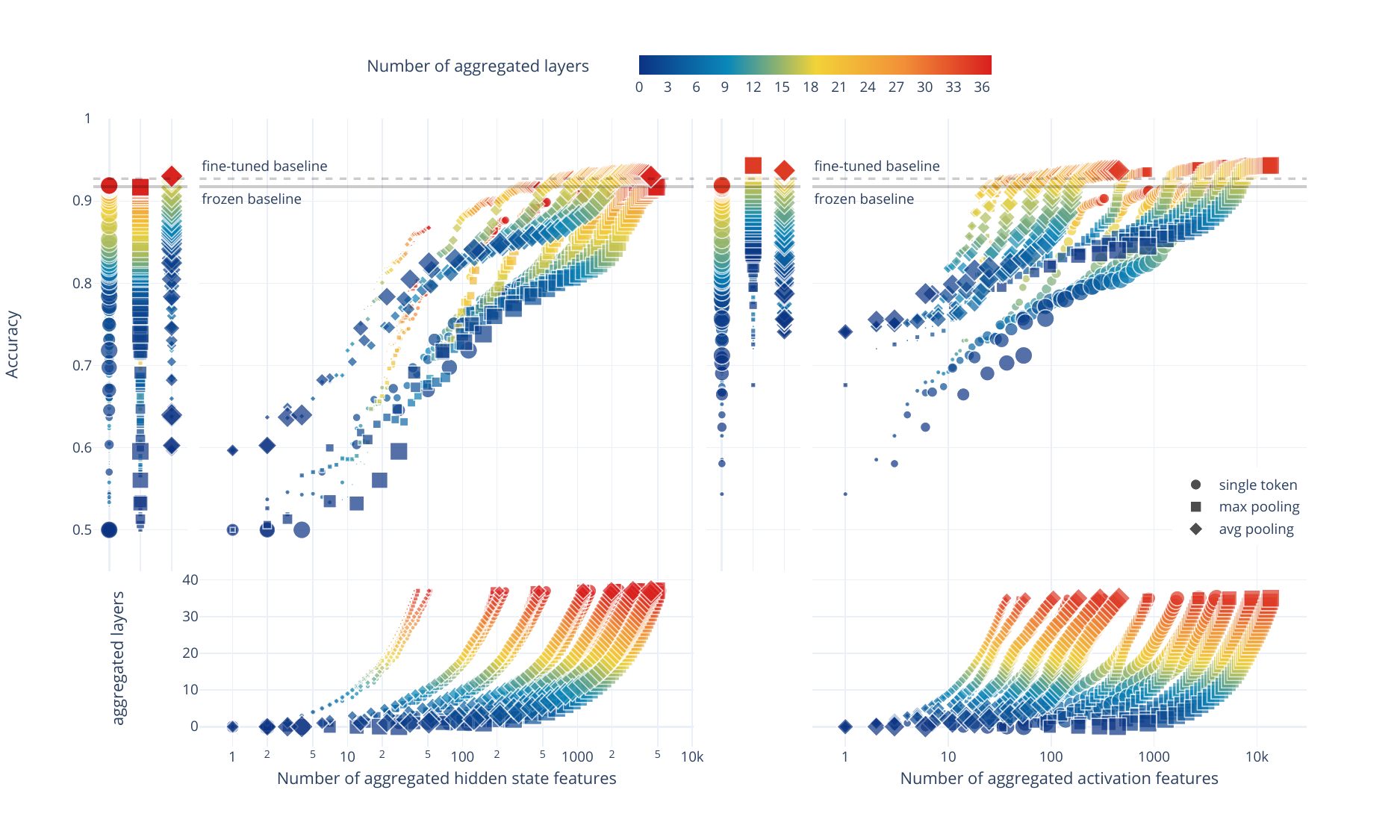}}
    \\
    \subfloat[Fine-tuned model]{
        \includegraphics[width=\textwidth]{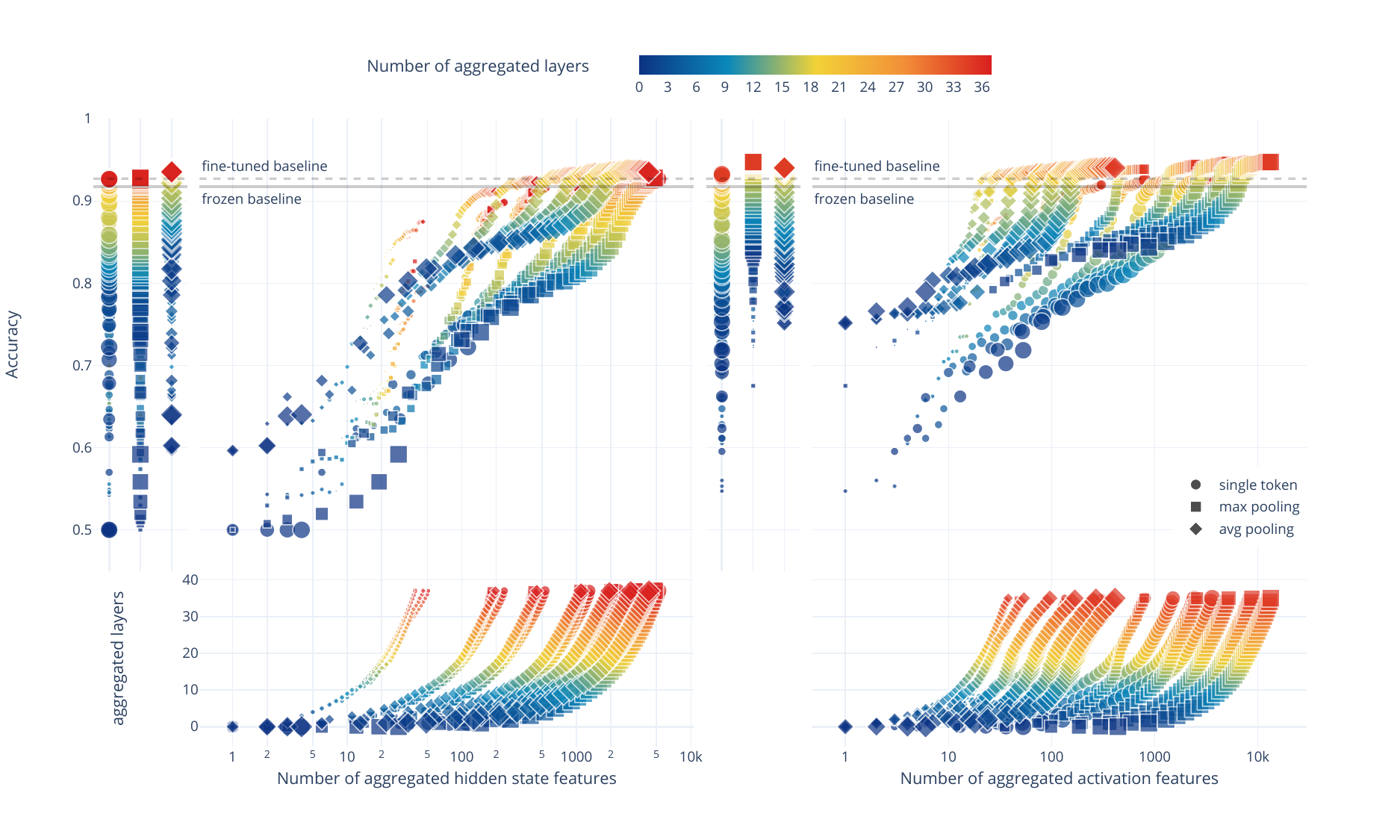}}
    \caption{What-which-where plots for the performance of \modelname{} on pretrained and fine-tuned GPT2-L models over IMDb dataset. The left subfigures are results with hidden states, and the right subfigures are with activations.}
\end{figure*}}

\begin{figure*}[htbp]
    \centering
    \subfloat[Pretrained model]{
        \includegraphics[width=\textwidth]{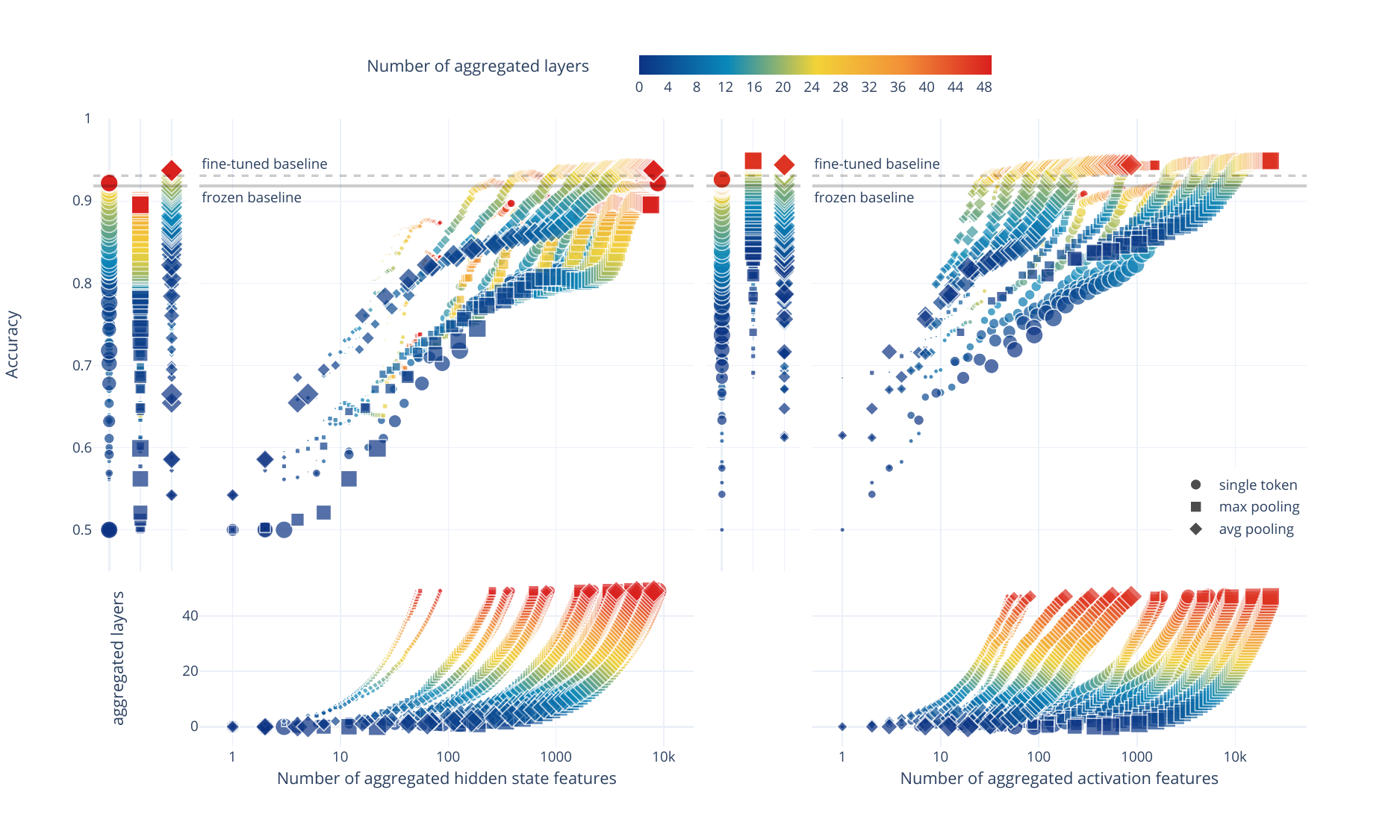}}
    \\
    \subfloat[Fine-tuned model]{
        \includegraphics[width=\textwidth]{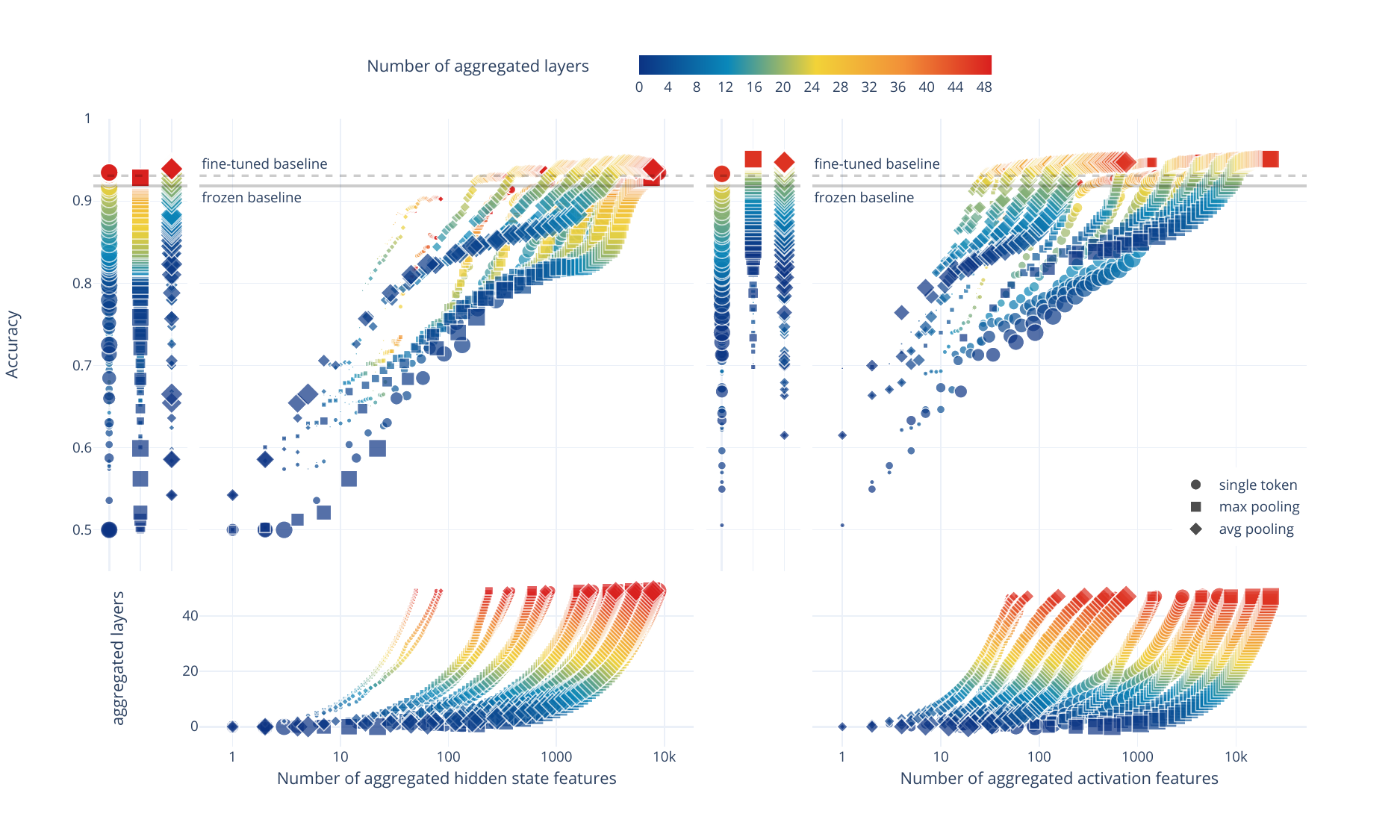}}
    \caption{What-which-where plots for the performance of \modelname{} on pretrained and fine-tuned GPT2-XL models over IMDb dataset. The left subfigures are results with hidden states, and the right subfigures are with activations.}
\end{figure*}

\end{document}